\documentclass[table, sigconf, nonacm]{acmart}
\AtBeginDocument{%
  }

\usepackage{xcolor}
\usepackage{makecell}
\usepackage{algorithm}
\usepackage{algpseudocode}
\usepackage{nicematrix}
\usepackage{hhline}
\usepackage{caption}
\usepackage{subcaption}
\usepackage{multirow}

\begin{document}

\title{A Post-Processing-Based Fair Federated Learning Framework}


\author{Yi Zhou}
\affiliation{%
  \institution{University of Oxford}
  \city{Oxford}
  \country{UK}}
\email{yi.zhou@keble.ox.ac.uk}

\author{Naman Goel}
\affiliation{%
  \institution{University of Oxford}
  \city{Oxford}
  \country{UK}}
\email{naman.goel@cs.ox.ac.uk}

\renewcommand{\shortauthors}{}

\begin{abstract}
Federated Learning (FL) allows collaborative model training among distributed parties without pooling local datasets at a central server. However, the distributed nature of FL poses challenges in training fair federated learning models. The existing techniques are often limited in offering fairness flexibility to clients and performance. We formally define and empirically analyze a simple and intuitive post-processing-based framework to improve group fairness in FL systems. This framework can be divided into two stages: a standard FL training stage followed by a completely decentralized local debiasing stage. In the first stage, a global model is trained without fairness constraints using a standard federated learning algorithm (e.g. FedAvg). In the second stage, each client applies fairness post-processing on the global model using their respective local dataset. This allows for customized fairness improvements based on clients' desired and context-guided fairness requirements. We demonstrate two well-established post-processing techniques in this framework: model output post-processing and final layer fine-tuning.
We evaluate the framework against three common baselines on four different datasets, including tabular, signal, and image data, each with varying levels of data heterogeneity across clients.
Our work shows that this framework not only simplifies fairness implementation in FL but also provides significant fairness improvements with minimal accuracy loss or even accuracy gain, across data modalities and machine learning methods, being especially effective in more heterogeneous settings.
\end{abstract}





\settopmatter{printfolios=true}
\maketitle
\section{Introduction}
In Federated learning (FL), multiple parties (clients) collaboratively train a model without sharing their local data~\cite{mcmahan_communication-efficient_2017}.
In addition to offering a way to use distributed data and compute for machine learning, federated learning is also investigated as a potential privacy-enhancing technology~\cite{zhao2024federated}. As a result, FL has attracted much attention~\cite{zhang2021survey}, with proposed applications in domains including healthcare, mobile devices and industrial engineering \cite{li2020review}.

Fair machine learning refer to a set of techniques to make machine learning models produce more fair outputs~\cite{barocas2023fairness}. For example, in some application contexts, the output of a binary classifier, that has been trained with historical data using machine learning, may need to satisfy certain fairness properties like conditional independence from attributes like gender and race~\cite{hardt_equality_2016,zafar_fairness_2017}. 

The literature in fair machine learning focuses mostly on centralized training setting, i.e. when all data used for training the model is located centrally. There has been recent work on understanding the challenges of fairness in federated learning setting as well. Federated learning inspired techniques train a global model with fairness constraints when data is distributed across multiple locations~\cite{zhang_fairfl_2020, du_fairness-aware_2020, abay_mitigating_2020, zeng_improving_2022}. Although these methods can be effective in improving fairness in FL, they also have a few limitations.
Firstly, these methods often lack fairness decentralization. 
Clients are required to agree on (a) the same fairness definition and the same fairness measurement, and (b) the same fairness requirements including the choice of sensitive attributes and protected groups, the level of fairness-accuracy trade-off, etc.
The requirement is problematic for several reasons.
One is that clients may have different needs depending on their specific use cases.
Another reason is that clients could have very different data distributions due to data heterogeneity in FL.
A single fairness constraint may not be suitable or work effectively on all clients, which may produce unpredictable model behavior on certain clients.
Secondly, many of these approaches require clients to use their statistics of sensitive attributes during distributed communication, which may raise additional privacy-related concerns.
Thirdly, imposing additional fairness-related constraints during global model training may also increase the network communication costs and slow down the training process, reducing the overall training efficiency.

To address these limitations, we explore a very simple post-processing-based fair FL framework.
In this framework, all clients first collaborate to perform global FL training without any fairness constraints.
Once the global training is finished, each client performs local fairness or debiasing post-processing on its own dataset,
tailoring the debiasing process towards its specific data distribution and requirements.
This method has a high flexibility and decentralization as it allows clients to apply different fairness constraints based on their specific needs, providing customized debiasing strategies.
It also addresses data heterogeneity since the debiasing is performed locally, it adapts fairness to the local data distribution.
Moreover, no information about sensitive attributes is required during the global training stage, thus offering better sensitive data protection.
This also improves training efficiency as there is no additional network communication cost for debiasing.

We demonstrate how post-processing debiasing methods from prior work can be integrated in this framework. In our experiments, we use two post-processing methods, but other methods can also be similarly integrated.
One is model output post-processing~\cite{hardt_equality_2016}.
This method addresses fairness by solving a linear program and computing a derived predictor. It does not need to change the original model weights and instead post-processes the model output. The other one is the final layer fine-tuning technique.
This method fine-tunes the last layer of neural network based models, with fairness constraints added to optimization objective. But it may be computationally more expensive and modifies model weights. While the term `post-processing' in fairness literature typically refers to model output post-processing methods, we use the term more broadly to include any kind of fairness post-processing, applied either to model output or model itself.

For evaluation, we conducted experiments to compare performance with three baseline methods: FedAvg~\cite{mcmahan_communication-efficient_2017}, FairFed~\cite{ezzeldin_fairfed_2023} and FairFed combined with Fair Linear Representation (FairFed/FR)~\cite{ezzeldin_fairfed_2023}. 
We included four different datasets in our experiments.
Two of them are commonly used tabular datasets in the fair ML literature, namely the Adult dataset~\cite{barry_becker_adult_1996} and the ProPublica COMPAS dataset~\cite{mattu_how_2016}.
The other two are from healthcare domain: a signal dataset, called PTB-XL Electrocardiogram (ECG) dataset~\cite{wagner_ptb-xl_2022}, and an image dataset, called NIH Chest X-Ray dataset~\cite{wang_chestx-ray8_2017}.
To simulate the different levels of data heterogeneity across clients, we use three heterogeneity settings for each dataset in the experiments to simulate the situations when the data partition is extremely imbalanced, medium-level imbalanced and slightly imbalanced.

Our experiments show that this framework consistently improves fairness across diverse data distributions. Moreover, both post-processing methods achieved these improvements with minimal computational costs, which makes the framework an efficient solution for real-world applications. A significant strength of this framework is also its simplicity and reliance on tried and tested algorithmic components from the literature. Finally, we also discuss the limitations of our work and opportunities for future work.

\section{Related Work} \label{sec:related}
In this section, we discuss prior work that are most closely related to this paper. For readers interested in a more comprehensive background in federated learning and machine learning fairness, we refer to recent surveys (such as~\cite{mehrabi_survey_2022, zhang2021survey}).

\paragraph{Federated Learning}
In federated learning (FL)~\cite{mcmahan_communication-efficient_2017}, different parties which are often called ``clients'' can collaborate with each other via a ``server'' without sharing their training data.
During each global training round of FL, each client computes gradient updates on its local dataset, and then, instead of sharing their data directly, they only share the updates with the server. The server aggregates the  updates from all the clients to compute an updated global model weights, and sends the new global modal to all clients for them to compute a new round of updates. This procedure repeats for a few global rounds until the global model reaches a certain performance or converges. Note that this process involves not only computation costs (for computing updates and aggregation), but also network communication costs (for sharing updates). Communication is often a bottleneck in distributed computation systems. 

One of the other major challenges in FL is that of data heterogeneity across clients \cite{mendieta_local_2022}.
In many real-world use cases of FL, different clients follow very different data distributions.
For example, in healthcare domain, different healthcare institutions might want to utilize all of their data to train a more accurate model for disease prediction. However, patient populations at different hospitals could have different feature distribution due to the location of hospitals, and a global model may not be better on all dimensions.

\paragraph{Personalized FL}
A sub-area in FL, known as Personalized FL (PFL) \cite{acar_debiasing_2021, wang_analyzing_2024}, is most closely related to our work. The overarching idea in PFL is to address data heterogeneity and individual needs by ``personalizing'' the global model towards a local objective.
\cite{li_ditto_2021} takes a slightly different approach to personalize the model by training local models and global models at the same time, and matching their similarity or merging them to a certain extent to improve the model performance while preserving some client-wise difference.
However, to the best of our knowledge, none of these works consider the fairness dimension.

\paragraph{Fairness in Machine Learning}
Due to the interdisciplinary nature of fairness in machine learning (and algorithmic decision-making systems broadly), a range of approaches and perspectives are necessary to understand the subject~\cite{holstein2019improving, barocas2023fairness}. We only focus on prior work that are most crucially relevant to understand the contribution of this paper. More specifically, we consider the relevant techniques for enforcing the group fairness notion of fairness~\cite{hardt_equality_2016, binns_apparent_2020}. While our framework can accommodate different fairness perspectives, in our experiments we mainly explore group fairness and leave experiments with other equally important fairness perspectives for future work.

Group fairness demands that different demographic groups (often distinguished by sensitive attributes such as gender and race) should have the same chance to receive certain (conditional) outcomes~\cite{hardt_equality_2016}. For e.g., conditioned on ``true'' qualification or label, group fairness may require a college admission decision classifier to ensure equal admission rates in different demographic groups. There are different versions of group fairness in the machine learning literature such as demographic parity, equal opportunity, equal odds, predictive value parity, accuracy parity, etc~\cite{barocas2023fairness}. The measure used in our experiments will be mathematically defined later. 

\paragraph{Fairness in FL}
In contrast to existing work for group fairness in FL~\cite{salazar2024survey}, our work decouples fairness implementation from global model training. The aim is to improve upon existing work by 1) decentralizing fairness i.e. letting the clients decide which fairness metrics, thresholds etc to implement in their local application context, 2) reducing computation and communication costs involved in global model training, and 3) relaxing the requirement of sensitive attribute data during global model training.

In FairFed\cite{ezzeldin_fairfed_2023}, clients send send their local updates as well as local fairness measurement to the server in each global round.
The server aggregates the local updates based on the fairness gap between each local fairness measurement and the global fairness (e.g. updates sent by ``fairer client'', i.e. client with the lowest fairness gap, have higher weight in global update).
All clients receive the same model satisfying one global notion of fairness after training.

Very recently, three more papers~\cite{che2024training,makhija2024achieving,duan2024post} have been submitted to arxiv (May, June and Nov 2024) that appear to share similarity with our work due to overlaps in some terms used in the papers, but on closer inspection we found that they are all fundamentally different from our work. \cite{che2024training} and ~\cite{makhija2024achieving} are closer to FairFed~\cite{ezzeldin_fairfed_2023} discussed above; they consider new ways of incorporating local fairness estimates during global training, but at the end of training all clients receive the same model satisfying one global notion of fairness. On the other hand, ~\cite{duan2024post} propose that clients use a post-processing function calculated by the server to achieve global fairness. In contrast to all these works, we completely decouple global model training (no fairness considerations) and local fairness interventions (no global coordination). Thus, we target different real-world scenarios/use-cases in which clients prefer the autonomy to decide fairness intervention independently.

A related line of work in fair federated learning (e.g. ~\cite{li_fair_2020,mohri_agnostic_2019,li_ditto_2021}) is about utility fairness for clients participating in FL. In contrast, we focus on the social attributes based fairness for decision subjects.

\section{Preliminaries}
\label{sec:prob_form}
We consider the following problem setting.
There are $K$ clients and one server participating in the FL training.
Each client $C_k$ ($k \in \{1,..,K\}$) has a local dataset $D_k$.
And all local datasets together form the global set $D = \cup_k D_k$. The distributions of $D_k$ may differ for different $k$, simulating heterogeneous data setting across different clients.

Our framework can be divided into mainly two stages, the training stage and the debiasing stage. 
Only the training stage requires message exchange between clients and the server, and the debiasing stage is performed fully locally, which means only communication rounds required for the basic FL framework are needed in our framework and there are no extra communication costs.

The training stage of this framework follows the training procedure of standard FL setting in FedAvg \cite{mcmahan_communication-efficient_2017}.
The general objective function in FedAvg can be written as follows (Equation \ref{eq:fedavg_obj}):

\begin{equation}
\begin{split}
\label{eq:fedavg_obj} f(\theta) =  \sum_{k=1}^{K} w_k l_k(\theta)
\end{split}
\end{equation}
where $\theta$ denotes the model parameters, $l_k$ denotes the local objective function on client $k$. The local objective function is fairness unaware, e.g. a vanilla loss function like cross-entropy loss.
Minimizing function $f(\theta)$ finds model parameter $\theta$ that minimizes the weighted average of the local model losses across all clients.

In the debiasing stage (which comes after the first stage is finished), our framework sends the global model to each client for them to evaluate the model locally on their dataset $D_k$, and apply different post-processing debiasing methods based on their local fairness constraint $F_k$.
In our experiments, we use Equalized Odds (EOD) as fairness metrics,
but note that due to the decoupling of global model training and debiasing in our framework, different clients are not required to follow the same fairness definition or constraint. Clients can adjust to different fairness metrics, different levels of fairness-accuracy trade-off, even different sensitive attributes based on their specific requirements. Clients can also use different post-process debiasing methods methods. Before discussing how we integrate post-process debiasing in this framework, we discuss EOD metric more formally.

Equalized Odds (EOD)~\cite{hardt_equality_2016} defines fairness as groups with different sensitive attributes having the same true positive rate (TPR) and false positive rate (FPR).
For evaluating and comparing EOD between different methods, we will measure EOD as the maximum of the absolute difference of TPR and the absolute difference of FPR between different groups, as shown in Equation \ref{eq:eod_global}. This is consistent with popular fairness toolkits like IBM's AI Fairness 360~\cite{bellamy_ai_2018}.

\begin{equation}
\begin{split}
\label{eq:eod_global}
    EOD = & Max(|TPR_{A=1} - TPR_{A=0}|, |FPR_{A=1} - FPR_{A=0}|) \\
    = & Max (|Pr(\hat{Y}=1|A=1, Y=1)- Pr(\hat{Y}=1|A=0, Y=1)|, \\
    & \qquad \quad |Pr(\hat{Y}=1|A=1, Y=0)- Pr(\hat{Y}=1|A=0, Y=0)|)
% \end{equation} 
\end{split}
\end{equation}
where $A$ is the sensitive attribute, ${Y}$ is the true label, and $\hat{Y}$ is the decision (or prediction).

For each client $C_k$, we measure local EOD metric as (Equation \ref{eq:eod_local}):

\begin{equation}
\begin{split}
\label{eq:eod_local}
    EOD_k = & Max(|TPR_{A=1, C_k} - TPR_{A=0, C_k}|, \\
    & \qquad \quad |FPR_{A=1, C_k} - FPR_{A=0, C_k}|) \\
    = & Max (|Pr(\hat{Y}=1|A=1, Y=1, C_k) \\
    & \qquad \qquad \qquad \quad - Pr(\hat{Y}=1|A=0, Y=1, C_k)|, \\
    & \qquad \quad  |Pr(\hat{Y}=1|A=1, Y=0, C_k) \\
    & \qquad \qquad \qquad \quad - Pr(\hat{Y}=1|A=0, Y=0, C_k)|)
\end{split}
\end{equation}
where $(\ |C_k)$ denotes that measure is calculated for client $C_k$ and their local data distribution (approximated using local dataset $D_k$).

\section{Post-Processing-Based Fair Federated Leaning Framework Details}
\label{sec:prop_method}
In this section, we present the post-processing-based framework with two example post-processing debiasing approaches: output post-processing method for any binary classifier~\cite{hardt_equality_2016} and final layer fine-tuning method for deep neural networks~\cite{mao_last-layer_2023}.
For both approaches, the global training process is based on the basic FedAvg training without fairness constraints as discussed above in Section \ref{sec:prob_form}. 
After the global training phase has finished, clients perform local post-processing methods on their copy of the global model.

\subsection{FL Model Output Fairness Post-Processing}
In Algorithm \ref{alg:pp}, we present the pseudo-code of our framework with FL model output fairness post-processing.

\begin{algorithm*}
\caption{FL Model Output Fairness Post-Processing}
\label{alg:pp}
\textbf{Initialize} server with global model weights $\omega_0$; $K$ clients with local training dataset $D_k$\hspace*{\fill}
\begin{algorithmic}[1]
\For{\texttt{each global round t=1, 2, ..., T}}
\Comment{FedAvg starts}
    \For{\texttt{each client k=1, 2, ..., K in parallel}}
        \State \texttt{$\omega_t^k$ $\gets$ ClientLocalUpdate($\omega_{t-1}, D_k$) }
        \Comment{Compute local update at each client using data $D_k$} \vspace{0.1cm}
        \State \texttt{CommunicateToServer($\omega_t^k$)}  \Comment{Communicate local update to server}
    \EndFor 
    \State \texttt{$\omega_t$ $\gets$ Aggregate$(\{\omega_{t}^k \}_{k=1}^K)$} \vspace{0.12cm}
    \Comment{Aggregate local updates to compute global update at server}
    \State \texttt{CommunicateToClients($\omega_t$)}
\EndFor 
% \Comment{FedAvg finished}
\For{\texttt{each client k=1, 2, ..., K in parallel}}
\Comment{Post-processing starts}
        \State \texttt{$\hat{Y}_k \gets$ Predict$(\omega_T, D_k)$}
        \Comment{Compute prediction with FedAvg model on local data $D_k$}
        \State \texttt{$p_k \gets $EqOdds$(\hat{Y}_k, D_k)$}
        \Comment{Compute derived predictor for client}
\EndFor
\end{algorithmic}
\end{algorithm*}

Algorithm \ref{alg:pp} starts by general FL training in lines 1-8~\cite{mcmahan_communication-efficient_2017}. At the termination of the global \texttt{for} loop, each client $k \in {1,..K}$ has received a trained FL model with weights $\omega_T$.
In line 10, predictions $\hat{Y_k}$ are computed using the global model $\omega_T$ on the local dataset $D_k$ for each client $C_k (k =1, 2, ... K)$.
In line 11, each client computes a derived predictor $p_k$ based on the prediction $\hat{Y_k}$ and the local dataset $D_k$. The method of obtaining the derived predictor is adopted from~\citet{hardt_equality_2016}. Before discussing the method of obtaining derived predictor, we stress that that each client computes a derived predictor separately based on their local context and requirements. For brevity, we do not distinguish between clients in the pseudo-code of Algorithm~\ref{alg:pp} (beyond the distinction in their datasets $D_k$). In principle, each client can also use different fairness definitions and post-processing methods in lines 10-11 and if a client doesn't wish to enforce fairness because of local application context, they can also skip lines 10-11.

\paragraph{Obtaining Derived Predictor~\cite{hardt_equality_2016}
} Formally, a derived predictor can be defined using the following probabilities:
$$p_{ya} = Pr(\tilde{Y} = 1 | \hat{Y}=y, {A}=a)$$
The four probabilities $p=(p_{00}, p_{01}, p_{10}, p_{11})$ specify the derived prediction $\tilde{Y}_p$. In other words, the derived predictor probabilistically flips the prediction of the model depending on the value of the sensitive attribute and the prediction, in order to optimize EOD fairness.

A client computes their local derived predictor $p$ using the following equation as the solution of a linear program:

\begin{equation}
\label{eq:loss_yy}
\begin{split}
    \min_{p} \textrm{ }  & \mathbb{E} \ {loss}(\tilde{Y}_{p}, Y) \\
    \textrm{s.t. } 
    & \gamma_0(\tilde{Y}_{p}) = \gamma_1(\tilde{Y}_{p}) \\
    & \forall_{y,a} \textrm{ } 0 \leq p_{ya} \leq 1 \\
\end{split}
\end{equation}
where 
% $Y$ is the sample label, 
$\hat{Y}$ is the prediction by the model, $\tilde{Y}_p$ is the prediction by the derived predictor $p$, $a$ is the value of sensitive attribute, and
\begin{equation*}
\label{eq:def_ry}
\begin{split}
    &\gamma_a(\tilde{Y}) \stackrel{\text{def}}{=\joinrel=} (Pr(\tilde{Y}=1 | A=a, Y=0), Pr(\tilde{Y}=1 | A=a, Y=1))
\end{split}
\end{equation*}

In other words, a client finds their derived predictor $p$ by minimizing the expected loss between the derived prediction $\tilde{Y}_p$ and true label $Y$ while satisfying the EOD fairness constraints $\gamma_0(\tilde{Y}_{p}) = \gamma_1(\tilde{Y}_{p})$. $Y$ and $A$ come from their local dataset $D_k$. 

\subsection{FL Model Final Layer Fairness Fine-Tuning}
In Algorithm~\ref{alg:ft}, we present the pseudo-code of our framework with FL model final layer fairness fine-tuning. 

Lines 1-8 are the same as Algorithm~\ref{alg:pp} since this is the global FedAvg model training stage. Each client receives a copy of the global model with weights $\omega_T$ at the end of this stage. In lines 9-15, we show the decoupled debiasing stage that is executed independently for each client. As was the case in the model output fairness post-processing method discussed previously, clients can decide their fairness definitions, metrics etc or skip fairness enforcement depending on their local context. The distinction between clients is not shown for brevity. On the other hand, unlike the model output fairness post-processing method, this method relies on modifying the weights in the final layer of a neural network model to improve fairness. \citet{mao_last-layer_2023} showed in centralized ML setting that fixing other model weights and only fine-tuning the last layer can effectively improve the fairness in neural networks.
%This method provides an efficient and inexpensive way of tuning especially large models like DNNs for fairness.
Besides, in FL, fine-tuning only the last layer can potentially be useful in preserving information learned through the data of other clients during global FL training.

In line 9, a client fixes the weights of the model layers except for the last layer $L$. Then $r$ rounds of fine-tuning are performed based on the local dataset $D_k$ on client $k$, with fine-tuning learning rate $\eta$ and the loss function $L$ that considers both local accuracy based loss ($l$) and local fairness $F_k$ dependent loss ($l'$). A parameter $\alpha$ assigns relative weight to the two losses. In our experiments, we use a loss $l'$ for EOD i.e. the sum of the differences in TPR and FPR for two demographic groups (see Section~\ref{sec:prob_form}).

\begin{algorithm*}
\caption{FL Model Final Layer Fairness Fine-Tuning}
\label{alg:ft}
% \hspace*{\algorithmicindent} \textbf{Input} \\
\textbf{Initialize} global model with $L$ layers and initial weights $\omega_0$; $K$ clients with local training dataset $D_k$; accuracy-based loss function $l$; fairness-based loss function $l'$; fine-tuning parameter $\alpha$; fairness metrics $F_k$ for each client $k$;
fine-tuning learning rate $\eta$ \hspace*{\fill}
% \textcolor{red}{l(): loss function ...; $\alpha$ parameter}
\begin{algorithmic}[1]
\For{\texttt{each global round t=1, 2, ..., T}}
\Comment{FedAvg starts}
    \For{\texttt{each client k=1, 2, ..., K in parallel}}
        \State \texttt{$\omega_t^k$ $\gets$ ClientLocalUpdate($\omega_{t-1}, D_k$) }
        \Comment{Compute local update at each client using data $D_k$} \vspace{0.1cm}
        \State \texttt{CommunicateToServer($\omega_t^k$)}  \Comment{Communicate local update to server}
    \EndFor 
    \State \texttt{$\omega_t$ $\gets$ Aggregate$(\{ \omega_{t}^k \}_{k=1}^K)$} \vspace{0.12cm}
    \Comment{Aggregate local updates to compute global update at server}
    \State \texttt{CommunicateToClients($\omega_t$)}
\EndFor 
% \Comment{FedAvg finished}
\For{\texttt{each client k=1, 2, ..., K in parallel}}
\Comment{Fine-tuning starts}
    \State \texttt{$\omega^{'k}\gets$ FreezeLayers$(\{\omega_T\}_{l=1}^{L-1}$})
    \Comment{Freeze weights for layer from 1 to (L-1)}
        \For{\texttt{each fine-tuning round r=1, 2, ..., R}}
            \State \texttt{ $L = \alpha$$l(\omega^{'k}, D_k) +  l'(\omega^{'k}, D_k, F_k) $ }
            \Comment{local weighted loss $L$ including fairness $F_k$}
            \State \texttt{ $\omega^{'k}$ $\gets$ $\omega^{'k} - \eta \nabla_{\omega^{'k}}(L)$ }
            \Comment{Local update at client}
        \EndFor 
\EndFor
\end{algorithmic}
\end{algorithm*}

\section{Experiment Settings}
\label{sec:dataset}
Next, we discuss the experiments settings for our empirical analysis of the the strengths and limitations of the framework using different real-world datasets. We also compare performance compare with different baselines.

\subsection{Datasets}\label{sec:data-datasets}
We first introduce the datasets used in our empirical analysis. We use two tabular datasets that are widely used in fair machine learning literature, namely Adult~\cite{barry_becker_adult_1996} and ProPublica COMPAS~\cite{mattu_how_2016}. In addition, we use two other datasets that are both larger in size and more complex in structure compared with the tabular datasets. One of them is an ECG signal dataset called PTB-XL~\cite{wagner_ptb-xl_2022}, and the other is a chest X-ray image dataset called NIH Chest X-Ray ~\cite{wang_chestx-ray8_2017}.

\paragraph{Adult Dataset}
Adult dataset \cite{barry_becker_adult_1996} is a popular dataset for individual's annual income prediction, and it is widely used in fairness evaluation of machine learning algorithms.
The target label in the Adult dataset is ``income'' which is divided into two classes (``<=50K'' and ``>50K'') for binary classification task.
We use ``sex'' as the sensitive attribute with ``male'' as 1 and ``female'' as 0 in our experiments.
We use one-hot encoding for all categorical features and apply \texttt{sklearn StandardScaler} \cite{pedregosa_scikit-learn_2011} to standardize continuous features.
The dataset originally consisted of 48842 samples. After dropping N/A values including values filled with abnormal values like `?', we have 44993 samples with 14662 (32.6\%) samples being ``female'' and 30331 (67.4\%) being ``male''.
Other statistics of the Adult dataset are shown in Table \ref{tab:adult}.

\begin{table}[h!]
\centering
\begin{tabular}{c | c | p {1cm} | p {1cm} | p {1cm} } 
\hline
 &  & \multicolumn{3}{c}{{$income$}}  \\
 \hline
 &   &  1 (>50K) & 0 (<=50K) & Total \\
\hline
  \multirow{5}{*}{\makecell{$sex$}}  
  & 1 (Male) & 9343 (20.8\%) & 20988 (46.6\%) & 30331 (67.4\%) \\
  & 0 (Female) & 1636 (3.6\%) & 13026 (29.0\%) & 14662 (32.6\%) \\
  \cline{2-5}
  & Total & 10979 (24.4\%)  & 34014 (75.6\%) & 44993 (100.0\%)\\
 \hline
 \hline
\end{tabular}
\caption{Statistics for the Adult dataset.  $income$ is used as target label, and $sex$ is used as sensitive attribute.}
\label{tab:adult}
\end{table}

\paragraph{COMPAS Dataset}
The ProPublica COMPAS dataset \cite{mattu_how_2016} is another popular dataset in machine learning fairness literature. Target variable in this datasets is observed recidivism within 2 years.
The dataset originally consisted of 7214 samples, and after dropping N/A values there are 6172 samples left.
We use $race$ as the sensitive attribute. 
For data pre-processing, we followed the procedures in \cite{mattu_how_2016} and \cite{zafar_fairness_2017}.
Samples with the race ``Caucasian'' are categorized as the privileged group and samples with the race ``African-American'' as the unprivileged group.
Samples in other race groups are dropped for the scope of experiments.
5278 samples are used for the experiments with 2103 samples ($39.8\%$) being ``Caucasian'' and 3175 samples being (60.2\%) ``African-American''.
We only use a subset of the data features including $``age\_cat"$, $``sex"$, $``priors\_count"$, $``c\_charge\_degree"$, as well as sensitive attribute $``race"$ and target label $``two\_year\_recid"$.
Similar to Adult data, we use one-hot encoding for categorical features and standardize numerical features using \texttt{sklearn StandardScaler}.
For detailed statistics of the COMPAS dataset, see Table \ref{tab:compas}.

\begin{table}[h!]
\centering
\begin{tabular}{c | c | p {1cm} | p {1cm} | p {1cm}} 
\hline
 &  & \multicolumn{3}{c}{{$two\_year\_recid$}}  \\
 \hline
 &  Total &  1 & 0 & Total \\
\hline
  \multirow{5}{*}{\makecell{$race$}}  
  & 1 (Caucasian) & 822 (15.5\%) & 1281 (24.3\%) & 2103 ($39.8\%$) \\
  & 0 (African-American) & 1661 (31.5\%) & 1514 (28.7\%) & 3175 (60.2\%) \\
  \cline{2-5}
  & Total & 2483 ($47.0\%$)  & 2795 ($53.0\%$) & 5278 (100.0\%)\\
 \hline
 \hline
\end{tabular}
\caption{Statistics for COMPAS dataset. $two\_year\_recid$ is used as target label, and $race$ is used as sensitive attribute.}
\label{tab:compas}
\end{table}

% \subsection{Signal/Image}

\begin{figure}[t]
\centering
\includegraphics[clip, trim=1.2cm 1cm 2.5cm 1cm, width=0.5\textwidth]{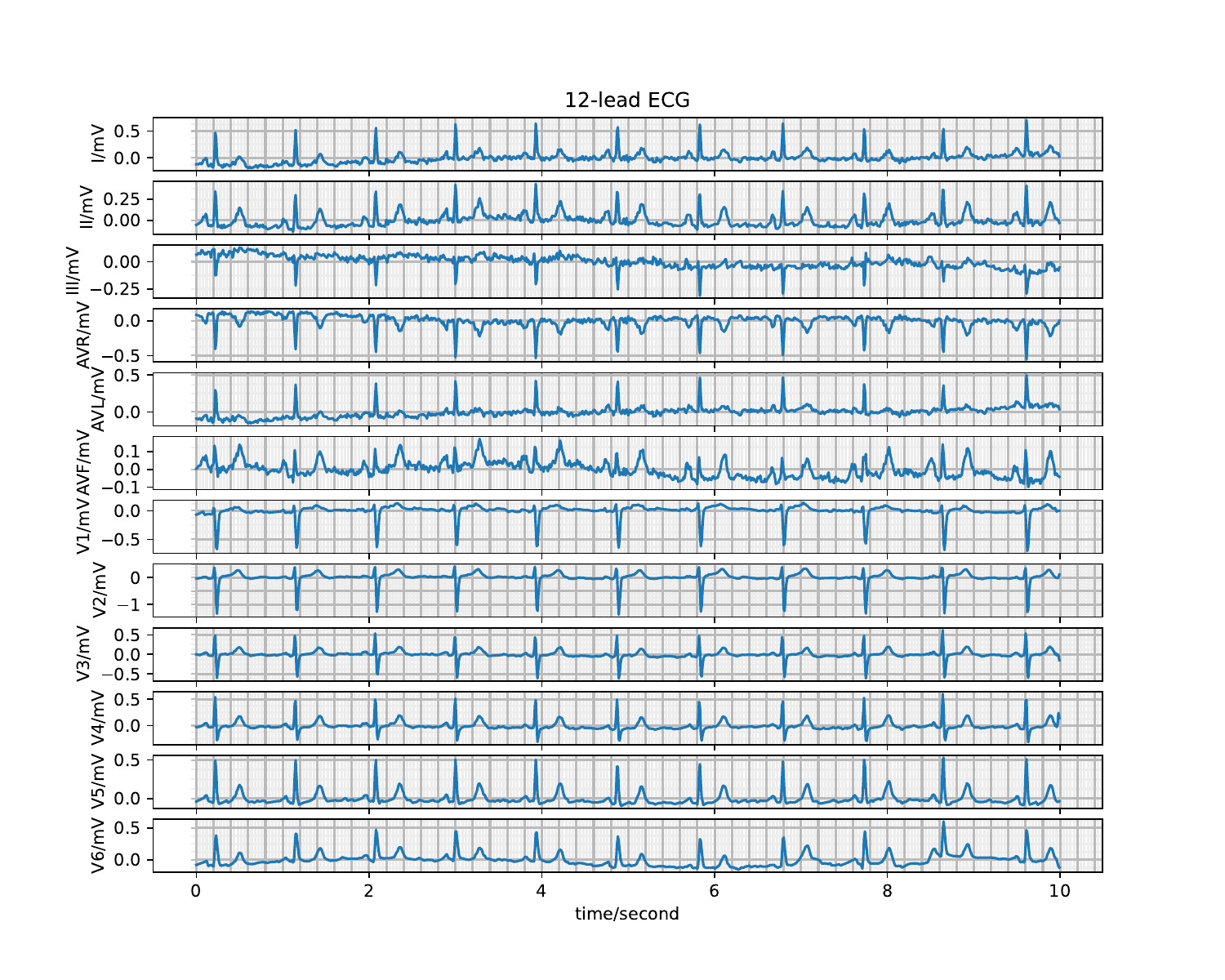} 
\caption[An example of ECG record from PTB-XL dataset]{A sample 12-lead ECG record from PTB-XL dataset. Y-axis: voltage of 12 different leads (channels). }
\label{fig:ecg_example}\end{figure}

\paragraph{PTB-XL ECG Signal Dataset}
The PTB-XL dataset \cite{wagner_ptb-xl_2022} is a large-scale electrocardiography (ECG) dataset consisting of 21799 clinical ECG signal records of 10-second length from 18869 patients.
The ECGs are 12-lead with a sampling frequency of 100Hz. Figure \ref{fig:ecg_example} shows an ECG sample from the dataset, with each row being the ECG signal from one ECG lead.
The 12 standard leads recorded in the dataset are lead I, II, III, aVF, aVR, aVL, V1, V2, V3, V4, V5, and V6, and each of them records the heart's electrical activity from a distinct viewpoint.
Together, they provide a multidimensional and comprehensive view of the heart's electrical activity, which makes ECG very commonly used and powerful tool in cardiology.

In the PTB-XL dataset, each ECG record is annotated by one or two cardiologists and assigned a statement to diagnose if the patient has a normal ECG or certain heart disease. The likelihoods for each diagnostic ECG statement are also provided.

For pre-processing, we dropped N/A values as well as entries with abnormal ages like ``300''.
Additionally, we only used samples with 100\% confidence diagnosis for the experiments.
We consider the target label as ``normal'' (ECG), and ECGs labeled with different heart diseases are considered as ``abnormal''.
We use patient age as the sensitive attribute and patients with age larger than 60 are classified into one group and the rest as the other group.
In the experiments, a total of 15766 samples are used with 5971 (37.9\%) classified as ``normal''.
Detailed statistics of PTB-XL are in Table \ref{tab:ptb}.

\begin{table}[h!]
\centering
\begin{tabular}{c | c | p{1.55cm} | p{1.55cm} | p{1.85cm} } 
\hline
 &  & \multicolumn{3}{c}{{$normal$}}  \\
 \hline
 &  Total &  1 & 0 & Total \\
\hline
  \multirow{3}{*}{\makecell{$age>60$}}  
  & 1  & 1987 (12.6\%) & 6717 (42.6\%) & 8704 ($55.2\%$) \\
  & 0  & 3984 (25.3\%) & 3078 (19.5\%) & 7062 (44.8\%) \\
  \cline{2-5}
  & Total & 5971 ($37.9\%$)  & 9795 ($62.1\%$) & 15766 (100.0\%)\\
 \hline
\end{tabular}
\caption{Statistics for PTB-XL dataset. $normal$ (ECG) is used as target label, and $age>60$ is used as sensitive attribute.}
\label{tab:ptb}
\end{table}

\begin{figure}
     \centering
     \begin{subfigure}[b]{0.23\textwidth}
         \centering
         \includegraphics[width=\textwidth]{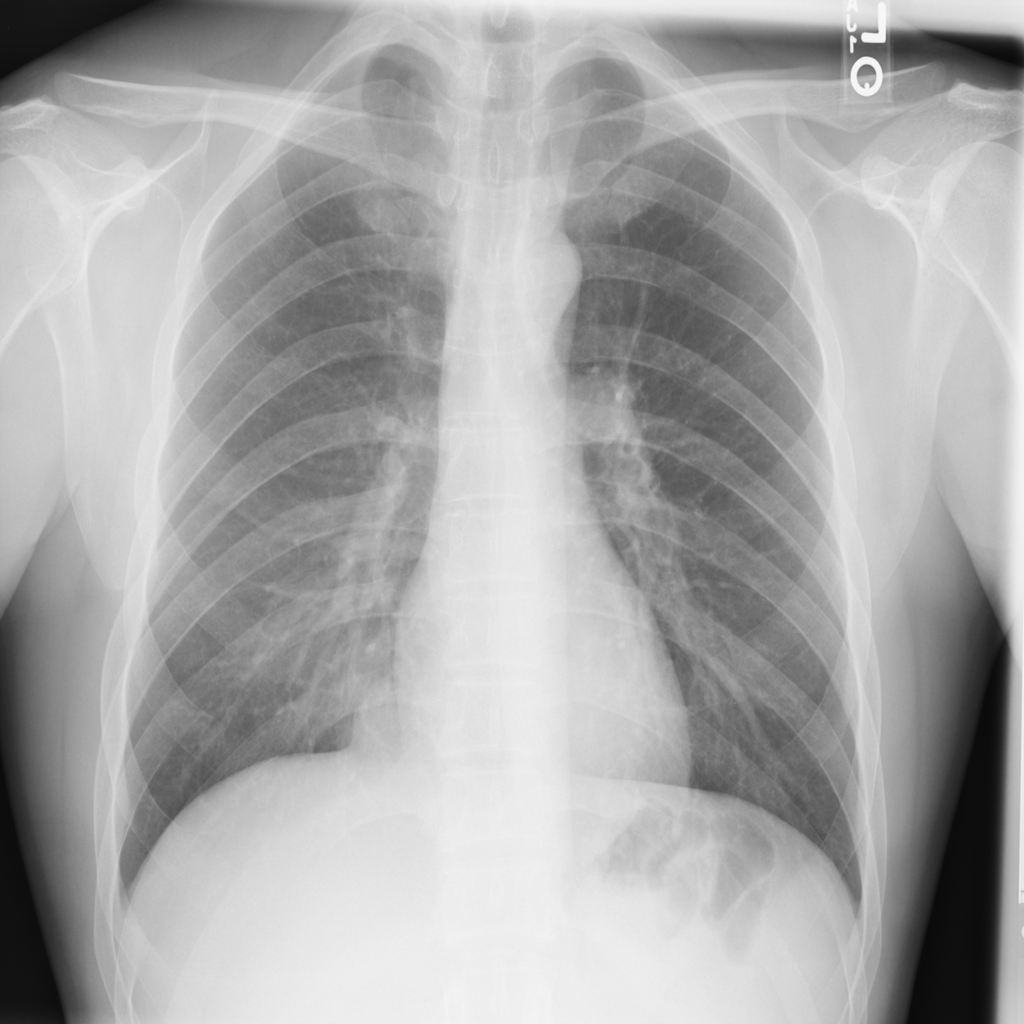}
     \end{subfigure}
     \hfill
     \begin{subfigure}[b]{0.23\textwidth}
         \centering
         \includegraphics[width=\textwidth]{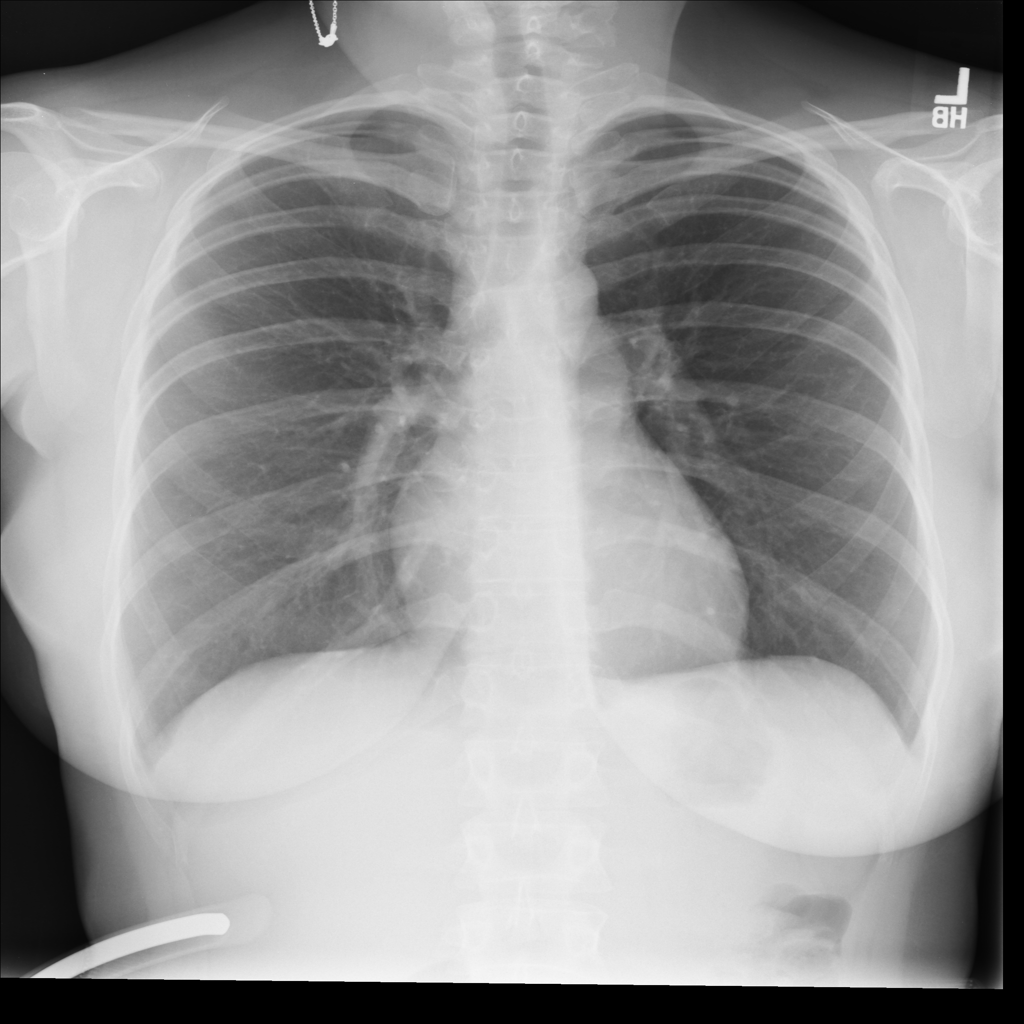}
     \end{subfigure}
     \hfill
     \begin{subfigure}[b]{0.23\textwidth}
         \centering
         \includegraphics[width=\textwidth]{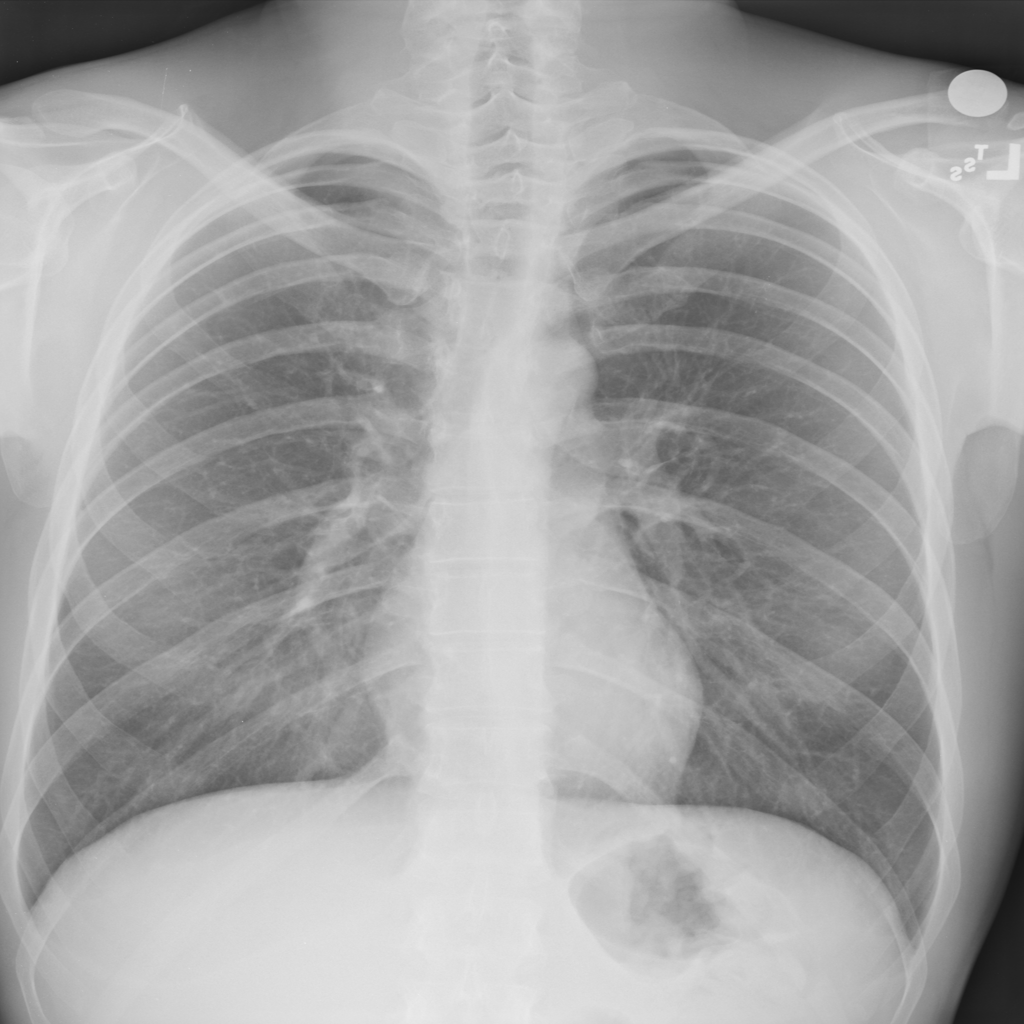}
     \end{subfigure}
     \hfill
     \begin{subfigure}[b]{0.23\textwidth}
         \centering
         \includegraphics[width=\textwidth]{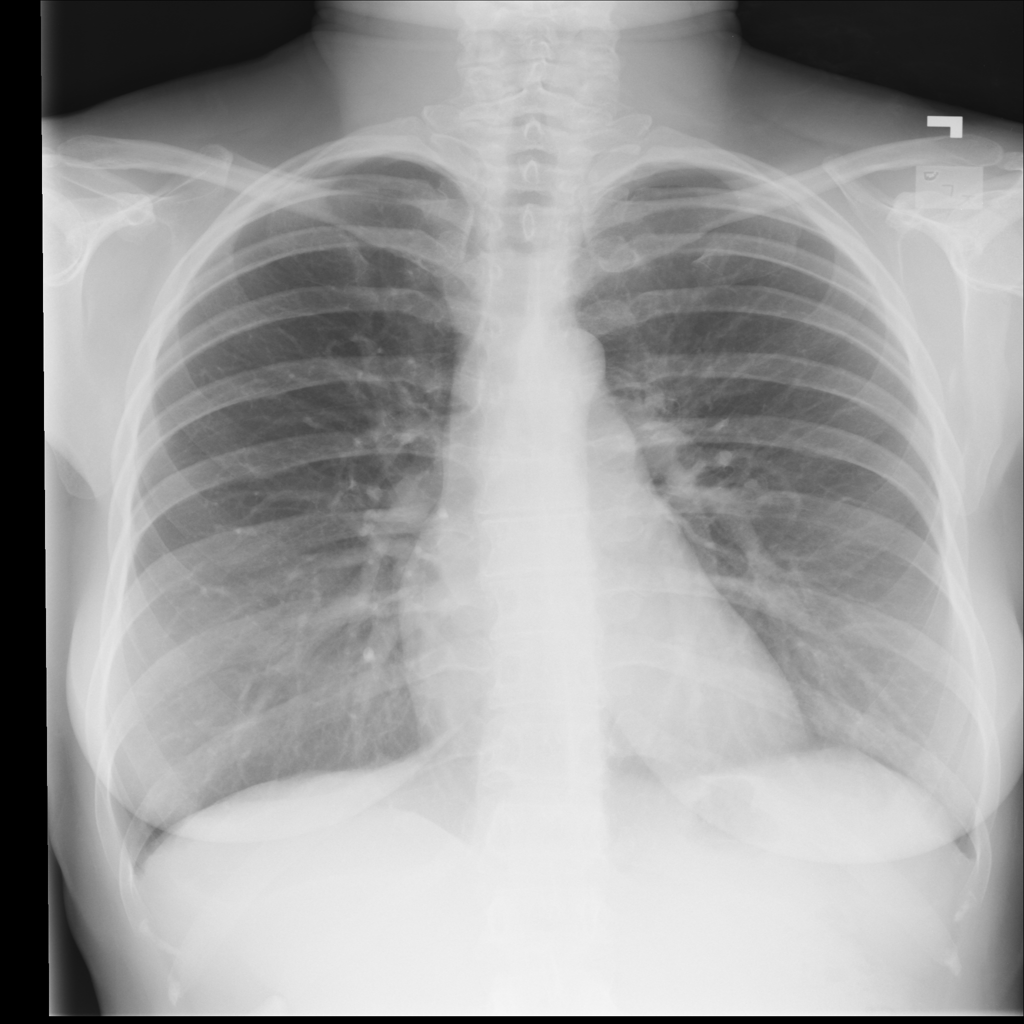}
     \end{subfigure}
        \caption{Four sample Chest X-ray images from NIH Chest X-Ray dataset.}
        \label{fig:nih_sample}
\end{figure}

\paragraph{NIH Chest X-Ray Image Dataset}
NIH Chest X-Ray dataset \cite{wang_chestx-ray8_2017} is a large-scale medical imaging dataset comprised of 112120 X-ray images from 30805 patients.
Figure \ref{fig:nih_sample} shows a few sample X-Ray images from the dataset.
The X-ray images come labeled with up to 14 diseases and ``No finding'' by natural language processing (NLP) models based on the original radiological reports of each X-ray. We only select samples with ``No Findings'' and disease ``Effusion'' for the scope of our experiments.
We use ``Effusion'' as the target label, and ``Patient gender'' as the sensitive attribute.
After removing entries filled with N/A and abnormal values, we have a dataset with 73669 samples, and 13316 (18.1\%) are labeled ``Effusion''.
In addition, we resized each image into size (256 * 256 * 3) with 3 channels both for computational reasons and the requirement for using pre-trained models during the training process (to be discussed in Section \ref{sec:hyper}).
Images are also normalized using the required mean and standard deviation based on the pre-trained model used in the experiments~\cite{sandler_mobilenetv2_2019}.  Detailed statistics of NIH Chest X-Ray are in Table \ref{tab:nih}.

\begin{table}[h!]
\centering
\begin{tabular}{c | c | p{1.7cm} | p{1.7cm}| p{1.85cm} }
\hline
\hline
 &  & \multicolumn{3}{c}{{$Effusion$}}  \\ [0.5ex] 
 \hline
 &  Total &  1 & 0 & Total \\
\hline
\hline
  \multirow{3}{*}{\makecell{$gender$}}  
  & 1  & 7434 (10.1\%) & 33916 (46.0\%) & 41350 ($56.1\%$) \\
  & 0  & 5882 (8.0\%) & 26437 (35.9\%) & 32319 (43.9\%) \\
  \cline{2-5}
  & Total & 13316 ($18.1\%$)  & 60353 ($81.9\%$) & 73669 (100.0\%)\\
 \hline
 \hline
\end{tabular}
\caption{Statistics for NIH-Chest X-Ray dataset. $Effusion$ is used as target label and $gender$ as sensitive attribute.}
\label{tab:nih}
\end{table}

Note that ``gender'' and ``sex'' are used interchangeably in the experiment datasets. 
In the NIH Chest X-Ray dataset, the feature $gender$ refers to the biological sex of patients.
Here we just kept the original feature name in the dataset, which is $gender$.

\paragraph{Remark on the use of ECG and chest X-Ray datasets} While prior works do consider similar health related datasets, fairness measures and sensitive attributes~\cite{marcinkevics_debiasing_2022, wang_analyzing_2024, grote2022enabling, noseworthy2020assessing, larrazabal2020gender}, ultimately the ethical considerations for fairness interventions are application and context dependent. Predicting abnormality or risk of mortality using health data is different from using such prediction in an application where fairness may be desired (e.g. for deciding access to a health intervention under resource constraints). While detailed application dependent discussions are beyond our expertise and the scope of this paper, we stress that such discussions with ethical, legal, domain experts and various stakeholders must be taken into account before employing any fairness intervention (including the ones that are examined in our paper). This further shows the utility of decoupling the two stages (federated model training and decentralized fairness post-processing by clients),  as proposed in this paper. By empirically analyzing our framework on ECG and chest X-Ray datasets, we only intend to examine framework's capability of dealing with complex data modalities and complex model architectures. Note that for Adult and COMPAS data too, various considerations about fairness interventions must be considered; albeit these datasets appear more frequently in the fairness literature compared to ECG and X-Ray datasets.

\subsection{Data Split}
To simulate the data distributions across different clients, we first partition the whole dataset to form local datasets without overlap. Then for each client, we split the local dataset into a training set and a test set, and only the training set is used during model development.

\subsection*{Clients Split}
We use a 4-client setup for all the datasets \cite{shamsian_personalized_2021}. To generate heterogeneity among clients, we use Dirichlet Distribution to sample data points, which is a commonly used in fair FL literature \cite{ezzeldin_fairfed_2023, shamsian_personalized_2021,yurochkin_bayesian_2019}.
We sample $p_i \sim Dir(\alpha)$ for each client $i \in \{1,..,4\}$, where $\alpha$ is a parameter controlling the level of data heterogeneity.
Smaller $\alpha$ provides a more heterogeneous distribution, and as $\alpha \rightarrow \infty$, the data distribution gets close to homogeneous and can be considered as random split~\cite{ezzeldin_fairfed_2023}.

To simulate different levels of data heterogeneity from extremely imbalanced to (very close to) random split, we start with $\alpha$ = 0.5, 1, 5, 100, 500  in Section \ref{sec:exp_heter}.
After comparing a method's performance under these different settings, we restrict to $\alpha$ = 0.5, 5, 500 for the following experiments since these appeared to be three broadly distinguishable heterogeneity levels.

Initially, we also considered $\alpha$ = 0.1 as one of the heterogeneity settings.
However, we found that due to the small dataset size and imbalanced label split, there would be missing labels within some groups with $\alpha$ = 0.1. 
For example, Table \ref{tab:diria01} shows an example data distribution on the COMPAS dataset generated from the Dirichlet distribution $Dir(\alpha)$ with $\alpha=0.1$.  We can see that the data partition is very imbalanced with only one sample or no samples for certain groups. This creates problems in our experiments (e.g. inability to even measure fairness on some clients).
Therefore, we consider $\alpha$ = 0.5 as the highest level of data heterogeneity for our experiment setting.
Similarly, we observed the data heterogeneity with $\alpha$ = 500 to be very small and quite close to random splits. Thus we will consider $\alpha$ = 500 as the lowest level of heterogeneity in the experiments.

\begin{table}[h!]
\centering
\begin{tabular}{c | c  c  c  c } 
\hline
\hline
Client & \#(Y=1,A=1) & \#(Y=1,A=0) & \#(Y=0,A=1) & \#(Y=0,A=0) \\
\hline
1 & 115 & 44 & 663 & 0 \\
2 & 233 & 86 & 1341 & 1 \\
3 & 7 & 582 & 1 & 691 \\
4 & 2 & 693 & 1 & 818 \\
 \hline
 \hline
\end{tabular}
\caption{An example data distribution with $Dir(\alpha=0.1)$ on COMPAS dataset. \#(Y=y, A=a) denotes the number of samples on a given client with label $y$ and sensitive attribute $a$. }
\label{tab:diria01}
\end{table}

\subsection*{Train/Test Split}
We split train and test sets on each client independently. On each client, we use 80\% samples as training set and 20\% as test set.
The global model is trained using all local training sets in the first stage (i.e. federated training). During the fairness post-processing and fine-tuning stage, we use local training set at the respective client.

\subsection{Baselines}\label{sec:baselines}
\paragraph{FedAvg: }
FedAvg refers to the originally proposed FL training framework without any fairness considerations~\cite{mcmahan_communication-efficient_2017} . Each client computes its local update and sends it to the server for aggregation and global model update.

\paragraph{FairFed: }
As discussed in Section~\ref{sec:related}, FairFed \cite{ezzeldin_fairfed_2023} is
based on the FedAvg framework but it uses a 
fairness-aware aggregation.

\paragraph{FairFed + Fair Representation (FairFed/FR): }

Fair Representation \cite{yuzi_he_geometric_2020} is a pre-processing method which debias the input features by removing their correlations with sensitive attributes.
We include FairFed combined with FairRep\cite{yuzi_he_geometric_2020} as one of our baselines, for the reason that the authors of the FairFed paper\cite{ezzeldin_fairfed_2023} presented the performance of FairFed combined with local pre-processing. Under their experimental setting, FairFed works well with Fair Representation with heterogeneous data distributions too.

\subsection{Evaluation Metrics}
We now describe the evaluation metrics we used in experiments for performance comparisons between different methods, following conventions in the literature.

\paragraph{Accuracy} 
We use the local (client-level) test accuracy as one of the main performance evaluation metrics on the tabular datasets Adult and COMPAS. It can be calculated using Equation \ref{eq:accuracy}:

\begin{equation}
\label{eq:accuracy}
\begin{split}
    \textrm{Accuracy} = \frac{\textrm{Number of correct predictions}}{\textrm{Number of total samples}}
    = \frac{\textrm{TP} + \textrm{TN}}{ \textrm{(TP+FN+FP+TN)} }
\end{split}
\end{equation}

\paragraph{Balanced Accuracy} 
For PTB-XL and NIH Chest X-Ray, we use balanced accuracy \cite{brodersen_balanced_2010} instead of accuracy for model performance evaluation. The reason is that health related datasets are often very imbalanced with labels for disease prediction or other classification tasks. In most cases, within a given dataset only a minority of patients is diagnosed with the specific disease, and the prediction accuracy can be high even if the model just predicts every sample as ``No disease''.
Balanced accuracy addresses this issue by including both sensitivity (TPR) and specificity (TNR) (Equation \ref{eq:ba}), providing a more reliable measurement for imbalanced datasets.
It is widely used in health ML literature such as \cite{maior_convolutional_2021, holste_long-tailed_2022}.

\begin{equation}
\label{eq:ba}
\begin{split}
    \textrm{Balanced Accuracy (BA)} = \frac{\textrm{Sensitivity (TPR)} + \textrm{Specificity (TNR)}}{2}
\end{split}
\end{equation}

\paragraph{EOD}
For fairness, we use local (client-level) Equalized Odds (EOD) as introduced in Section \ref{sec:prob_form}.
It is defined as the maximum of the absolute TPR difference between different groups and the absolute FPR difference between different groups (see Equation \ref{eq:eod_local}).

\paragraph{Weighted Average}
Although we focus on local metrics, we also include a weighted average of accuracy, balanced accuracy, and EOD. It can be interpreted as a measurement of the average performance across clients. We calculate the weighted average of a given measure of a metric $x$ by:

\begin{equation}
\label{eq:w_avg}
\begin{split}
    \textrm{Weighted Average} = \sum_{k=1}^{K} (\frac{n_k  x_k}{\sum_{k=1}^{K} n_k})
\end{split}
\end{equation}
where $n_k$ denotes the number of samples on client $k$ and $x_k$ denotes the local measure of $x$ on client $k$.

\paragraph{(Total Training) Time}
The total model training time includes both the FL training time and the debiasing time.
Although for our framework the FL training time and the debiasing time can be measured separately, this is not feasible for the FairFed baselines.
In FairFed debiasing procedure happens within the FL training process.
Therefore, we will use total time for training and debiasing (in baseline as well as in our framework) for fair performance comparison.

\paragraph{Communication Rounds}
Communication rounds measure the number of message exchanges between the server and clients.
Every time when the server receives a message from a client and a client receives a message from the server will be counted as one communication round.  

\section{Model Development} \label{sec:hyper}

\subsection{Model Architecture}
Adult and COMPAS tabular datasets are generally evaluated with simple architectures in the literature~\cite{zafar_fairness_2017}. We use
a simple one-layer model to train on the Adult and COMPAS datasets for our methods and all the baselines.
The activation function ReLU \cite{agarap_deep_2019} is used in the model. We use stochastic gradient descent (SGD) \cite{ruder_overview_2017} as the optimizer and Binary Cross Entropy Loss (BCELoss) \cite{mao_cross-entropy_2023} as the loss function when training on Adult and COMPAS.

For the ECG dataset PTB-XL, we used a ResNet-based model \cite{lima_deep_2021} with residual blocks \cite{he_identity_2016} to handle the uni-dimensional ECG signals. The model is composed of one convolutional layer and five residual blocks. Each residual block consists of two convolutional layers by batch normalization \cite{ioffe_batch_2015}, activation function ReLU \cite{agarap_deep_2019} and Dropout regularizer \cite{srivastava_dropout_2014}.
We use Adam optimizer \cite{kingma_adam_2017} with weighted mean square error loss function for the ECG dataset. 

In the case of NIH Chest X-Ray dataset, we initialize our model with the pre-trained model MobileNetV2 \cite{sandler_mobilenetv2_2019}. The MobileNetV2 model is trained on Imagenet. Prior works such as \cite{reshan_detection_2023} have shown that utilizing pre-trained models (including MobileNet) for initialization is useful for classification tasks on X-Ray datasets.

\paragraph{Code Availability} To supplement the above information, we also provide our implementation/code in the supplementary material. Code will be available on GitHub \url{https://github.com/Yi-Zhou-01/fairpostprocess-fedml}. As we will discuss in the next subsection, all algorithms require hyperparameters to be tuned. For FairFed and FairFed/FR, we report results with hyperparameters settings based on the information available in the paper and that showed best results for these baselines in our implementation.

\subsection{Hyperparameters}
The hyperparameters we used for methods on different datasets can be found in Table \ref{tab:hyperparameters}.
Note that although we only listed three methods in the table (FedAvg, FairFed and FT), all the methods used in the experiments are covered.
The FL training stage of our methods PP (output post-processing) and FT (fine-tuning) are the same as FedAvg, and PP does not require hyperparameter tuning.
FairFed/FR uses the same parameters as FairFed since Fair Representation is a pre-processing method and does not require extra hyperparameter tuning.

\begin{table}[ht]
\centering
\begin{tabular}{c | c c | c c c | c c c}
 \hline
 \multirow{2}{*}{ Dataset } & \multicolumn{2}{c|}{FedAvg} & \multicolumn{3}{c|}{FairFed } &
 \multicolumn{3}{c}{FT } 
 \\
& lr & bs & lr & bs & $\beta$ & lr & bs & $\alpha_{ft}$\\ [0.5ex] 
 \hline
 Adult & 0.01 & 32 & 0.01 & 32 & 0.1 & 5e-3 & 256 & 1.0\\ 
 COMPAS & 0.01 & 32 & 0.01 & 32 & 0.5 & 5e-3 & 256 & 2.0 \\
 PTB-XL & 5e-3 & 32 & 5e-3 & 32 & 0.1 & 5e-3 & 512 & 1.0\\
 NIH-Chest & 1e-4 & 64 & 1e-4 & 64 & 0.1 & 1e-4 & 64 & 0.1 \\
 \hline
\end{tabular}
\caption{Hyperparameters used in the experiments for different methods on all the datasets. 
lr: learning rate. bs: local batch size. $\beta$: fairness budget in FairFed. $\alpha_{ft}$: fine-tuning parameter in FT method. Both of our methods use the same parameters as FedAvg during the FL training stage. FT denotes the final-layer fine-tuning stage of our method.}
\label{tab:hyperparameters}
\end{table}

\subsection{Libraries and Computational Resources}
We use Python as the programming language and PyTorch \cite{paszke_pytorch_2019} for the machine learning model development and experiments. We also use IBM AIF 360 \cite{bellamy_ai_2018} for the output post-processing approach and for fairness evaluation. Besides, libraries including sklearn, numpy, pandas, Pillow \cite{noauthor_python_2024}, and H5py are also used in the experiments for data processing and model training. For training on signal and image datasets, we use NVIDIA Tesla V100 for GPU acceleration.

\section{\label{ch:results} Results and Discussion}

\subsection{Experiment with Different Heterogeneity Levels}\label{sec:exp_heter}

\begin{figure}[ht]
\centering\includegraphics[width=0.5\textwidth]{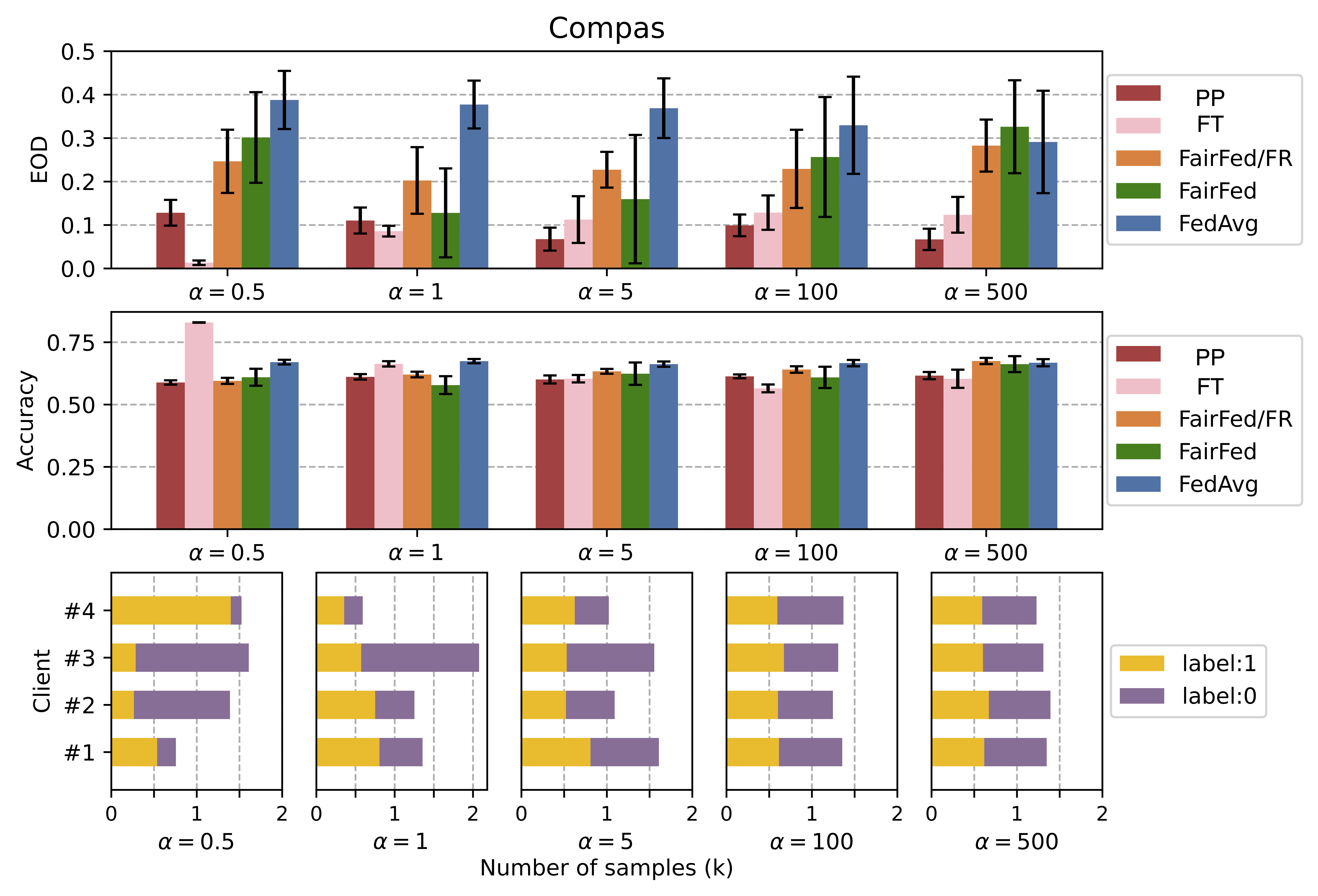} 
\caption[Experiment with different data heterogeneity levels using COMPAS dataset]{Experiment using COMPAS dataset with different data heterogeneity level  $\alpha$. Smaller $\alpha$ indicates more heterogeneous data partitions. 
Accuracy and EOD are weighted averaged across clients based on the dataset size; standard deviation (std) is shown with error bars. 1st row: Test accuracy (higher value is better). 2nd row: Equalized odds difference (lower value is better). 3rd row: Sample distribution on each client with different labels.
FedAvg, FairFed, and FairFed/FR are baselines described in Section \ref{sec:baselines}. PP and FT refer to our framework with output post-processing and final layer fine-tuning respectively.}
\label{fig:plot_compas_alpha}
\end{figure}

In this experiment, we first explore the impact of data heterogeneity level on the performance of different methods. We partition the COMPAS dataset across four clients using Dirichlet distribution $Dir(\alpha)$ with $\alpha= 0.5, 1, 5, 100, 500$, and evaluate the performance of different methods.
The third row in Figure \ref{fig:plot_compas_alpha} shows the distribution of data labels across clients under different heterogeneity settings. 
As expected, lower values of $\alpha$ (e.g., $\alpha = 0.5$ and $\alpha = 1$) result in more imbalanced data partitions both in terms of label distribution and dataset size. In contrast, higher values ($\alpha = 100$ and $\alpha = 500$) generate more balanced splits, with $\alpha = 500$ closely simulating an identically distributed setting, mimicking random data splits across clients. The medium level of heterogeneity is represented by $\alpha = 5$ under our experimental settings.

The first row of Figure \ref{fig:plot_compas_alpha} shows the EOD values for each method as data heterogeneity increases. Experiments were repeated for 10 random seeds under each setting and the average and standard deviation are reported. 
We can observe that firstly, the EOD for the standard FL model without fairness constraints (FedAvg) decreases as $\alpha$ increases, which suggest that more balanced splits generally result in less bias (lower EOD values are better). As for FairFed and FairFed/FR, the best performance is observed at around medium to high levels of data heterogeneity ($\alpha=0.5, 1$). However, we can see that the performance of FairFed and FairFed/FR is not very robust with a large standard deviation (std). A possible reason could be sensitivity w.r.t. the FairFed hyper-parameter $\beta$.

On the other hand, our framework with both PP and FT methods shows a clear improvement in EOD under all heterogeneity settings with much smaller standard deviations. Simple output post-processing method can achieve the highest improvement when the data is not too imbalanced ($\alpha= 5,100,500$).
When the heterogeneity is extremely high ($\alpha= 0.5,1$), final-layer fine-tuning shows its effectiveness by reducing EOD close to zero.

The second row of Figure \ref{fig:plot_compas_alpha} provides the test accuracy with respect to data heterogeneity.
Generally, models trained without any fairness constraints would obtain a higher accuracy compared with fairness fairness-constrained models. This situation can be observed in almost every method in Figure \ref{fig:plot_compas_alpha}.
However, when $\alpha=0.5$, we can see that FT even achieves an increase in accuracy with a significant EOD improvement.
The reason could be that, the fine-tuning procedure helps the model to optimize towards both the local fairness and accuracy goal as shown in line 12 of Algorithm~\ref{alg:ft}, addressing data heterogeneity problem in fairness and prediction accuracy at the same time.

Henceforth, we will use three different data heterogeneity settings with $\alpha=0.5, 5$ and $500$ to simulate
high, medium and very low (close to homogeneity) levels of data heterogeneity respectively.
Further, note that the final-layer fine tuning (FT) method is meant for deep neural networks. For the sake of completeness of analysis in Figure~\ref{fig:plot_compas_alpha}, we used a 2-layer neural network for COMPAS dataset, but in all further discussions, we will use a simple one-layer network for both tabular datasets Adult and COMPAS (as explained previously in Section~\ref{sec:hyper}). This also means that we will not discuss the FT method for tabular datasets henceforth. For brevity reasons, we have prioritized reporting results that are most practically relevant rather than overloading the readers with unnecessary numbers.

\subsection{Performance on COMPAS Dataset}
\begin{table}[htbp]
\centering
\begin{tabular}{c | c | c | c c c c} 
 % \hline
 % \hline
 \multicolumn{7}{c}{\textbf{COMPAS Dataset}} \\ [0.5ex] 
 \hline
 \hline 
  % \multirow{2}{*}{\makecell{Hetero.\\Level $\alpha$}}
  \multirow{2}{*}{\makecell{$\alpha$}} 
  & \multirow{2}{*}{Metric} & \multirow{2}{*}{Client} &  \multicolumn{4}{c}{Method}\\
 \cline{4-7}
 % \hline
 & &
 & FedAvg & FairFed & FF+FR & PP \\
 % \cline{2-8}
 \hline
 \hline
  % \multirow{11}{*}{$\alpha=0.5$} 
  \multirow{11}{*}{$0.5$}  
  & \multirow{5}{*}{Acc.} 
  &  Avg & 0.670 & \cellcolor[gray]{0.9} \textbf{0.674} & 0.652 & 0.610 \\
  && C1 & 0.632 & \cellcolor[gray]{0.9} \textbf{0.705} & 0.628 & 0.550 \\
  && C2 & 0.633 & \cellcolor[gray]{0.9} \textbf{0.658} & 0.638 & 0.532 \\
  && C3 & \cellcolor[gray]{0.9} \textbf{0.729} & 0.705 & 0.684 & 0.706 \\
  && C4 & \cellcolor[gray]{0.9} \textbf{0.704} & 0.682 & 0.656 & 0.697  \\
    
  % \cline{2-8}
  \hhline{~|*{6}{-}}
  & \multirow{5}{*}{EOD} 
  &  Avg & 0.418 & 0.378 & 0.407 & \cellcolor[gray]{0.9} \textbf{0.088} \\
  && C1 & 0.464 & 0.405 & 0.415 & \cellcolor[gray]{0.9} \textbf{0.078} \\
  && C2 & 0.370 & 0.348 & 0.379 & \cellcolor[gray]{0.9} \textbf{0.064} \\
  && C3 & 0.387 & 0.340 & 0.376 & \cellcolor[gray]{0.9} \textbf{0.101} \\
  && C4 & 0.532 & 0.484 & 0.524 & \cellcolor[gray]{0.9} \textbf{0.137}  \\

  % \cline{2-8}
  \hhline{~|*{6}{-}}
  & \multicolumn{2}{c|}{Time} 
  & \cellcolor[gray]{0.9} \textbf{7 sec} & 8 sec & 8 sec & \cellcolor[gray]{0.9}\textbf{7 sec} \\
  \hhline{~|*{6}{-}}
  & \multicolumn{2}{c|}{\makecell{Comm. Rounds}} 
  & \cellcolor[gray]{0.9} \textbf{324} & \cellcolor[gray]{0.9} \textbf{324} & \cellcolor[gray]{0.9} \textbf{324} & \cellcolor[gray]{0.9} \textbf{324} \\
  \hline
  \hline
  % \multirow{11}{*}{$\alpha=5$} 
  \multirow{11}{*}{$5$} 
  & \multirow{5}{*}{Acc.} 
  & Avg & \cellcolor[gray]{0.9}\textbf{0.670} & 0.663 & 0.660 & 0.602 \\
  && C1 & \cellcolor[gray]{0.9}\textbf{0.671} & 0.661 & 0.661 & 0.583 \\
  && C2 & {0.648} & 0.640 & \cellcolor[gray]{0.9}\textbf{0.654} & 0.591 \\
  && C3 & {0.708} & \cellcolor[gray]{0.9}\textbf{0.712} & 0.703 & 0.679 \\
  && C4 & \cellcolor[gray]{0.9}\textbf{0.636} & 0.613 & 0.599 & 0.527\\
  
  % \cline{2-8}
  \hhline{~|*{6}{-}}
  & \multirow{5}{*}{EOD} 
  & Avg & 0.392 & 0.317 & 0.334 & \cellcolor[gray]{0.9}{\textbf{0.086}} \\
  && C1 & 0.454 & 0.368 & 0.369 & \cellcolor[gray]{0.9}{\textbf{0.062}} \\
  && C2 & 0.345 & 0.271 & 0.312 & \cellcolor[gray]{0.9}{\textbf{0.118}} \\
  && C3 & 0.356 & 0.293 & 0.329 & \cellcolor[gray]{0.9}{\textbf{0.075}} \\
  && C4 & 0.402 & 0.312 & 0.312 & \cellcolor[gray]{0.9}{\textbf{0.106}} \\
  
  % \cline{2-8}
  \hhline{~|*{6}{-}}
  & \multicolumn{2}{c|}{Time}  & \cellcolor[gray]{0.9}{\textbf{7 sec}} & 11 sec & 11 sec & \cellcolor[gray]{0.9}{\textbf{7 sec}}  \\
    \hhline{~|*{6}{-}}
  & \multicolumn{2}{c|}{\makecell{Comm. Rounds}} 
  & \cellcolor[gray]{0.9} \textbf{324} & \cellcolor[gray]{0.9} \textbf{324} & \cellcolor[gray]{0.9} \textbf{324} & \cellcolor[gray]{0.9} \textbf{324} \\
  \hline
  \hline
  % \multirow{11}{*}{$\alpha=500$}
  \multirow{11}{*}{$500$} 
  & \multirow{5}{*}{Acc.} 
  &  Avg     & \cellcolor[gray]{0.9}\textbf{0.673} & 0.671 & 0.672 & 0.603 \\
  && C1 & {0.671} & 0.666 & \cellcolor[gray]{0.9}\textbf{0.676} & 0.602 \\
  && C2 & \cellcolor[gray]{0.9}\textbf{0.680} & {0.679} & 0.678 & 0.607  \\
  && C3 & \cellcolor[gray]{0.9}\textbf{0.665} & 0.661 & 0.664 & 0.611 \\
  && C4 & {0.675} & \cellcolor[gray]{0.9}\textbf{0.678} & {0.668} & 0.589\\

  % \cline{2-8}
  \hhline{~|*{6}{-}}
  & \multirow{5}{*}{EOD} 
  & Avg & 0.416 & 0.414 & 0.374 & \cellcolor[gray]{0.9}\textbf{0.064} \\
  && C1 & 0.422 & 0.434 & 0.391 & \cellcolor[gray]{0.9}\textbf{0.053} \\
  && C2 & 0.401 & 0.368 & 0.336 & \cellcolor[gray]{0.9}\textbf{0.061} \\
  && C3 & 0.399 & 0.424 & 0.400 & \cellcolor[gray]{0.9}\textbf{0.080} \\
  && C4 & 0.444 & 0.434 & 0.373 & \cellcolor[gray]{0.9}\textbf{0.061}\\

  % \cline{2-8}
  \hhline{~|*{6}{-}}
  & \multicolumn{2}{c|}{Time}  & \cellcolor[gray]{0.9}\textbf{7 sec} & 11 sec & 11 sec & \cellcolor[gray]{0.9}\textbf{7 sec} \\
    \hhline{~|*{6}{-}}
  & \multicolumn{2}{c|}{\makecell{Comm. Rounds}} 
  & \cellcolor[gray]{0.9} \textbf{324} & \cellcolor[gray]{0.9} \textbf{324} & \cellcolor[gray]{0.9} \textbf{324} & \cellcolor[gray]{0.9} \textbf{324} \\
  \hline
  \hline
\end{tabular}
\caption{Client-wise performance comparison on COMPAS dataset with different data heterogeneity level $\alpha$. Smaller $\alpha$ indicates more imbalanced partitions. 
Experiments are repeated for 10 random seeds under each setting and the average of 10 runs are presented. Acc.: Test accuracy (higher is better). EOD: Equalized odds difference (lower is better). 
Grey highlights indicate the best performance in each row.
FedAvg, FairFed, and FF+FR are baselines described in Section \ref{sec:baselines}. 
PP refers to our framework with model output post-processing.}
\label{tab:stats_compas}
\end{table}

In Table \ref{tab:stats_compas}, we present the client-wise performance of our framework with output post-processing (PP) method compared to the baselines, under varying data heterogeneity. 
In general, our framework with PP shows a significant improvement in fairness at all clients as well as in the weighted average of local fairness under all heterogeneity settings. In contrast, the FairFed and FairFed/FR baselines show lower fairness improvements. The performance of the baselines in also less consistent, which suggests the sensitivity of FairFed-based approaches to experimental conditions. However, the improvement in fairness in PP comes at a cost of relatively more drop in accuracy.

\paragraph{Performance under extreme heterogeneity}
Under extreme heterogeneity setting ($\alpha=0.5$), our framework with PP, outperforms all the baselines in fairness improvement across clients and also show a significant improvement in weighted averaged EOD ($\sim$ 79\% improvement over FedAvg).

\paragraph{Performance under medium to low heterogeneity}
When the data partitions are less heterogeneous (with $\alpha=5$ and $\alpha=500$), PP becomes slightly more effective in improving fairness with an improvement of $\sim$ 85\% for  $\alpha=500$ over FedAvg.

\paragraph{Training cost}
Table~\ref{tab:stats_compas} shows that our framework with PP have the same number of communication rounds as the basic FL framework FedAvg. This is because the FL training stage of our method is adopted from FedAvg and the debiasing stage is performed locally and does not need message exchange.

The FairFed-based baselines also happen to have the same number of communication rounds as FedAvg for each global round of training in this dataset, but as we will see in the case of other datasets, this is not always guaranteed. The communication cost per global round is also in theory higher in baselines because fairness related information is also exchanged in these rounds.

We finally note the total training time of baselines and our framework. We report the time after rounding and do not show decimal points for brevity. As we can see that our framework with PP has almost the same time as the FedAvg (i.e. fairness comes in our framework with PP at almost no time cost). On the other hand, the time is higher for FairFed based baselines. Due to computational resource (i.e. research budget) constraints, we simulated all the clients on a single machine instead of having different clients located in different locations with network latency between them. In real-world, the actual communication costs and training time would be higher due to the network transition costs and the differences between different methods would be more stark. 

The PP stage of our framework can be performed in parallel on each client as it is entirely local method. So the actual PP time should be the maximum of PP time across clients.
In our results, we report the sum of the PP time on all clients, which serves as a very generous upper bound of PP time for fair comparison.

\subsection{Performance on Adult Dataset}
\label{sec:adult}

\begin{table}[htbp]
\centering
\begin{tabular}{c | c | c | c c c c} 
 % \hline
 % \hline
 \multicolumn{7}{c}{\textbf{Adult Dataset}} \\ [0.5ex] 
 \hline
 \hline 
  % \multirow{2}{*}{\makecell{Hetero.\\Level $\alpha$}}
  \multirow{2}{*}{\makecell{$\alpha$}}
  & \multirow{2}{*}{Metric} & \multirow{2}{*}{Client} &  \multicolumn{4}{c}{Method}\\
 \cline{4-7}
 % \hline
 & &
 & FedAvg & FairFed & FF+FR & PP \\
 % \cline{2-8}
 \hline
 \hline
  % \multirow{11}{*}{$\alpha=0.5$} 
  \multirow{11}{*}{$0.5$} 
  & \multirow{5}{*}{Acc.} 
  & Avg &  \cellcolor[gray]{0.9} \textbf{0.847} & 0.833 & 0.784 & 0.821 \\
  && C1 & \cellcolor[gray]{0.9} \textbf{0.804} &  {0.767} & {0.741} & 0.774 \\
  && C2 & \cellcolor[gray]{0.9} \textbf{0.739} &  {0.704} & {0.648} & 0.694\\
  && C3 & 0.925 & \cellcolor[gray]{0.9} \textbf{0.958} & {0.924} & 0.908 \\
  && C4 & \cellcolor[gray]{0.9} \textbf{0.725} & 0.625 & 0.500 & 0.684 \\
  % \cline{2-8}

  \hhline{~|*{6}{-}}
  & \multirow{5}{*}{EOD} 
  &  Avg & 0.193 & 0.158 & 0.199 & \cellcolor[gray]{0.9} \textbf{0.060} \\
  && C1 & 0.154 & {0.108} & {0.122} & \cellcolor[gray]{0.9} \textbf{0.051} \\
  && C2 & 0.324 & {0.295} & \cellcolor[gray]{0.9} \textbf{0.118} & {0.214}  \\
  && C3 & 0.224 & 0.205 & 0.299 & \cellcolor[gray]{0.9} \textbf{0.068} \\
  && C4 & 0.171 & 0.110 & {0.076} & \cellcolor[gray]{0.9} \textbf{0.042}  \\

  % \cline{2-8}
  \hhline{~|*{6}{-}}
  & \multicolumn{2}{c|}{Time} 
  & \cellcolor[gray]{0.9} \textbf{30 sec} & 78 sec & 78 sec & \cellcolor[gray]{0.9}\textbf{30 sec} \\
  \cline{2-7}
  \hhline{~|*{6}{-}}
  & \multicolumn{2}{c|}{\makecell{Comm. Rounds}} 
  & \cellcolor[gray]{0.9} \textbf{164} & 244 & 244 & \cellcolor[gray]{0.9} \textbf{164} \\
  \hline
  \hline
  % \multirow{11}{*}{$\alpha=5$} 
  \multirow{11}{*}{$5$} 
  & \multirow{5}{*}{Acc.} 
  & Avg & \cellcolor[gray]{0.9} \textbf{0.842} & \cellcolor[gray]{0.9} \textbf{0.842} & 0.829 & 0.810  \\
  && C1 & \cellcolor[gray]{0.9} \textbf{0.784} & 0.780 & 0.757 & 0.743  \\
  && C2 & \cellcolor[gray]{0.9} \textbf{0.797} & 0.792 & 0.760 & 0.750 \\
  && C3 & 0.875 & \cellcolor[gray]{0.9} \textbf{0.879} & 0.873 & 0.852 \\
  && C4 & 0.872 & \cellcolor[gray]{0.9} \textbf{0.874} & 0.872 & 0.845\\

  % \cline{2-8}
  \hhline{~|*{6}{-}}
  & \multirow{5}{*}{EOD} 
  & Avg & 0.164 & 0.149 & 0.153 &  \cellcolor[gray]{0.9} \textbf{0.066} \\
  && C1 & 0.197 & 0.186 & 0.173 &  \cellcolor[gray]{0.9} \textbf{0.052} \\
  && C2 & 0.122 & 0.108 & 0.107 &  \cellcolor[gray]{0.9} \textbf{0.062} \\
  && C3 & 0.164 & 0.140 & 0.177 &  \cellcolor[gray]{0.9} \textbf{0.101} \\
  && C4 & 0.171 & 0.164 & 0.149 &  \cellcolor[gray]{0.9} \textbf{0.042} \\
  
  % \cline{2-8}
  \hhline{~|*{6}{-}}
  & \multicolumn{2}{c|}{Time}  & \cellcolor[gray]{0.9}{\textbf{30 sec}} & 78 sec & 78 sec & \cellcolor[gray]{0.9}{\textbf{30 sec}} \\
  \hhline{~|*{6}{-}}
  & \multicolumn{2}{c|}{\makecell{Comm. Rounds}} 
  & \cellcolor[gray]{0.9} \textbf{164} & 244 & 244 & \cellcolor[gray]{0.9} \textbf{164} \\
  \hline
  \hline
  % \multirow{11}{*}{$\alpha=500$} 
  \multirow{11}{*}{$500$} 
  & \multirow{5}{*}{Acc.} 
  &  Avg     & {0.841} & \cellcolor[gray]{0.9}\textbf{0.842} & {0.841} & 0.810 \\
  && C1 & {0.836} & \cellcolor[gray]{0.9}\textbf{0.838} & {0.835} & 0.806\\
  && C2 & {0.845} & \cellcolor[gray]{0.9}\textbf{0.846} & \cellcolor[gray]{0.9}\textbf{0.846} & 0.816 \\
  && C3 & \cellcolor[gray]{0.9}\textbf{0.842} & {0.841} & \cellcolor[gray]{0.9}\textbf{0.842} & 0.806 \\
  && C4 & {0.840} & \cellcolor[gray]{0.9}\textbf{0.841} & {0.840} & 0.812 \\

  % \cline{2-8}
  \hhline{~|*{6}{-}}
  & \multirow{5}{*}{EOD} 
  & Avg & 0.173 & 0.159 & 0.160 & \cellcolor[gray]{0.9}\textbf{0.044}\\
  && C1 & 0.178 & 0.152 & 0.152 & \cellcolor[gray]{0.9}\textbf{0.032} \\
  && C2 & 0.175 & 0.167 & 0.163 & \cellcolor[gray]{0.9}\textbf{0.031}\\
  && C3 & 0.144 & 0.147 & 0.148 & \cellcolor[gray]{0.9}\textbf{0.032}  \\
  && C4 & 0.195 & 0.171 & 0.177 & \cellcolor[gray]{0.9}\textbf{0.081} \\
  
  % \cline{2-8}
  \hhline{~|*{6}{-}}
  & \multicolumn{2}{c|}{Time}  & \cellcolor[gray]{0.9}\textbf{30 sec} & 78 sec & 78 sec & \cellcolor[gray]{0.9}\textbf{30 sec} \\
  \hhline{~|*{6}{-}}
  & \multicolumn{2}{c|}{\makecell{Comm. Rounds}} 
   & \cellcolor[gray]{0.9} \textbf{164} & 244 & 244 & \cellcolor[gray]{0.9} \textbf{164} \\
  % & \multicolumn{2}{c|}{\makecell{Communication\\Rounds}} & - & - & - & - \\
  \hline
  \hline
\end{tabular}
\caption{Client-wise performance comparison on Adult dataset with different data heterogeneity level $\alpha$, where smaller $\alpha$ indicates more imbalanced partition. 
Acc.: Test accuracy (Higher is better). EOD: Equalized odds difference (Lower is better). 
Grey highlights indicate the best performance in each row.
FedAvg, FairFed, and FF+FR are baselines described in Section \ref{sec:baselines}. 
PP refers to our framework with output post-processing.}
\label{tab:stats_adult}
\end{table}

Compared with the COMPAS dataset, the Adult dataset is larger in size but more imbalanced in terms of label and sensitive attribute distribution with only 3.6\% of samples having $label=1$ and $sensitive\_attribute=0$ (Table \ref{tab:adult}). Table \ref{tab:adult_example} shows a label distribution with $\alpha=0.5$ on Adult dataset for clients.

As shown in Table~\ref{tab:stats_adult}, our framework with PP shows its effectiveness in improving fairness under all heterogeneity and outperforms all baselines on Adult too.

\begin{table}[h!]
\centering
\begin{tabular}{c | p{1.3cm}  p{1.3cm}  p{1.3cm} p{1.45cm} | p{1cm} } 
\hline
C & \#(Y=1,A=1) & \#(Y=1,A=0) & \#(Y=0,A=1) & \#(Y=0,A=0) & Total \\
\hline
1 & 183 (12\%) & 486 (31\%) & 41 (3\%) & 870 (55\%) & 1580 (100\%)\\
2 & 699 (11\%) & 2039 (33\%) & 123 (2\%) & 3309 (54\%) & 6170 (100\%)\\
3 & 1141 (5\%) & 7651 (36\%) & 192 (1\%) & 12142 (57\%) & 21126 (100\%) \\
4 & 5459 (77\%) & 280 (4\%) & 945 (13\%) & 433 (6\%) & 6170 (100\%)\\
 \hline
\end{tabular}
\caption{An example local training data distribution with $Dir(\alpha=0.5)$ on Adult dataset.  C denotes different clients. \#(Y=y, A=a) denotes the number of samples on a given client with label $y$ and sensitive attribute $a$. }
\label{tab:adult_example}
\end{table}

\paragraph{Performance under different heterogeneity level}
Compared with the baselines, PP achieves the highest fairness improvement under all heterogeneity settings
with some accuracy decrease.
FairFed provides a much smaller EOD improvement while maintaining a similar accuracy level as FedAvg.
FairFed/FR generally appears to perform worse in terms of fairness-accuracy trade-off.

\paragraph{Training cost}
A larger difference in training time between our methods and FairFed-based baselines can be found in this experiment compared with the COMPAS experiment, mainly because of the larger dataset size of Adult dataset. The time spent on the post-processing stage in PP is still negligible resulting in almost the same running time after rounding as FedAvg.
FT method is also efficient with less than half of the training time of FairFed method.
In contrast, FairFed takes two to three times the running time as it needs to calculate a new global fairness metric and metric gap between global and local fairness metrics.
In addition, we note that our method require a smaller number of communication rounds. As discussed previously recall that communication cost per communication round too is smaller for our method compared to FairFed based baselines.

\begin{table}[htbp]
\centering
\begin{tabular}{c | c | c |c c c c} 
 % \hline
 % \hline
 \multicolumn{7}{c}{\textbf{PTB-XL Dataset}} \\ [0.5ex] 
 \hline
 \hline 
  % \multirow{2}{*}{\makecell{Hetero.\\Level $\alpha$}}
  \multirow{2}{*}{\makecell{$\alpha$}}
  & \multirow{2}{*}{Metric} & \multirow{2}{*}{Client} &  \multicolumn{4}{c}{Method}\\
 \cline{4-7}
 & &
 & FedAvg & FairFed & PP & FT \\
 \hline
 \hline
  % \multirow{11}{*}{$\alpha=0.5$} 
  \multirow{11}{*}{$0.5$} 
  & \multirow{5}{*}{BA} 
  &  Avg & 0.790 & 0.780 & 0.684 & \cellcolor[gray]{0.9}\textbf{0.803}  \\
  && C1 & 0.760 & \cellcolor[gray]{0.9}\textbf{0.773} & 0.642 &  0.771 \\
  && C2 & 0.810 & \cellcolor[gray]{0.9}\textbf{0.811} & 0.712 &  0.809\\
  && C3 & 0.734 & 0.767 & 0.581 & \cellcolor[gray]{0.9} \textbf{0.806}\\
  && C4 & \cellcolor[gray]{0.9}\textbf{0.858} & 0.841 & 0.800 & 0.853 \\
  \hhline{~|*{6}{-}}
  & \multirow{5}{*}{EOD} 
  & Avg & 0.342 & 0.308 & \cellcolor[gray]{0.9}\textbf{0.044} & {0.299} \\
  && C1 & 0.334 & 0.286 & \cellcolor[gray]{0.9}\textbf{0.074} & {0.308}\\
  && C2 & 0.369 & 0.330 & \cellcolor[gray]{0.9}\textbf{0.074} & {0.407} \\
  && C3 & 0.349 & 0.316 & \cellcolor[gray]{0.9}\textbf{0.021} &  {0.267}\\
  && C4 & 0.172 & 0.198 & \cellcolor[gray]{0.9}\textbf{0.080} & {0.186} \\
  \hhline{~|*{6}{-}}
  & \multicolumn{2}{c|}{Time} 
  & \cellcolor[gray]{0.9} \textbf{356 sec} & 416 sec & \cellcolor[gray]{0.9}\textbf{356 sec} & 391 sec \\
  \hhline{~|*{6}{-}}
  & \multicolumn{2}{c|}{\makecell{Comm. Rounds}} 
   & {324} & \cellcolor[gray]{0.9} \textbf{164} & 324 & {324} \\
  \hline
  \hline
  % \multirow{11}{*}{$\alpha=5$} 
  \multirow{11}{*}{$5$} 
  & \multirow{5}{*}{BA} 
  & Avg & 0.770 & 0.769 & 0.648 & \cellcolor[gray]{0.9}\textbf{0.790} \\
  && C1 & 0.791 & 0.787 & 0.672 & \cellcolor[gray]{0.9}\textbf{0.799} \\
  && C2 & 0.774 & \cellcolor[gray]{0.9}\textbf{0.778} & 0.675 & 0.776\\
  && C3 & 0.771 & 0.768 & 0.636 & \cellcolor[gray]{0.9}\textbf{0.777}\\
  && C4 & 0.755 & 0.750 & 0.615 & \cellcolor[gray]{0.9}\textbf{0.805}\\
  \hhline{~|*{6}{-}}
  & \multirow{5}{*}{EOD} 
  & Avg & 0.342 & 0.358 & \cellcolor[gray]{0.9}\textbf{0.055} & 0.319 \\
  && C1 & 0.412 & 0.431 & \cellcolor[gray]{0.9}\textbf{0.066} & 0.421\\
  && C2 & 0.347 & 0.342 & \cellcolor[gray]{0.9}\textbf{0.065} & 0.327\\
  && C3 & 0.360 & 0.392 & \cellcolor[gray]{0.9}\textbf{0.051} & 0.341\\
  && C4 & 0.288 & 0.311 & \cellcolor[gray]{0.9}\textbf{0.042} & 0.239\\
  \hhline{~|*{6}{-}}
  & \multicolumn{2}{c|}{Time}  & \cellcolor[gray]{0.9}{\textbf{154 sec}} & 224 sec & \cellcolor[gray]{0.9}{\textbf{154 sec}} & 174 sec  \\
  \hhline{~|*{6}{-}}
  & \multicolumn{2}{c|}{\makecell{Comm. Rounds}} 
   & \cellcolor[gray]{0.9} \textbf{164} & \cellcolor[gray]{0.9} \textbf{164} & \cellcolor[gray]{0.9} \textbf{164} & \cellcolor[gray]{0.9} \textbf{164} \\
  \hline
  \hline
  % \multirow{11}{*}{$\alpha=500$} 
  \multirow{11}{*}{$500$} 
  & \multirow{5}{*}{BA} 
  & Avg & {0.770} & {0.772} & 0.645 & \cellcolor[gray]{0.9}\textbf{0.793} \\
  && C1 & {0.754} & {0.753} & 0.629 & \cellcolor[gray]{0.9}\textbf{0.784} \\
  && C2 & {0.760} & {0.770} & 0.630 & \cellcolor[gray]{0.9}\textbf{0.789}  \\
  && C3 & {0.777} & 0.774 & 0.648 & \cellcolor[gray]{0.9}\textbf{0.797} \\
  && C4 & {0.788} & 0.791 & 0.671 & \cellcolor[gray]{0.9}\textbf{0.803} \\
  \hhline{~|*{6}{-}}
  & \multirow{5}{*}{EOD} 
  & Avg & 0.345 & 0.343 & \cellcolor[gray]{0.9}\textbf{0.046}  & 0.319 \\
  && C1 & 0.364 & 0.359 & \cellcolor[gray]{0.9}\textbf{0.029} & 0.334 \\
  && C2 & 0.335 & 0.334 & \cellcolor[gray]{0.9}\textbf{0.049} & 0.307 \\
  && C3 & 0.355 & 0.350 & \cellcolor[gray]{0.9}\textbf{0.047} & 0.340 \\
  && C4 & 0.328 & 0.332 & \cellcolor[gray]{0.9}\textbf{0.055} & 0.298 \\
  \hhline{~|*{6}{-}}
  & \multicolumn{2}{c|}{Time}  & \cellcolor[gray]{0.9}\textbf{151 sec} & 220 sec & \cellcolor[gray]{0.9}\textbf{151 sec} & 170 sec \\
  \hhline{~|*{6}{-}}
  & \multicolumn{2}{c|}{\makecell{Comm. Rounds}} 
   & \cellcolor[gray]{0.9} \textbf{164} & \cellcolor[gray]{0.9} \textbf{164} & \cellcolor[gray]{0.9} \textbf{164} & \cellcolor[gray]{0.9} \textbf{164} \\
  \hline
  \hline
\end{tabular}
\caption{Client-wise performance comparison on PTB-XL ECG dataset with different data heterogeneity levels $\alpha$.
BA: Balanced accuracy (Higher is better). EOD: Equalized odds difference (Lower is better). 
Grey highlights indicate the best performance of each row.
FedAvg and FairFed are baselines described in Section \ref{sec:baselines}. 
PP and FT refer to our framework with output post-processing and final-layer fine-tuning respectively.}
\label{tab:stats_ptb}
\end{table}

\subsection{Performance on PTB-XL ECG Dataset}
\label{sec:ptb}
Table \ref{tab:stats_ptb} presents the performance comparison of different methods on PTB-XL ECG dataset under different heterogeneity settings.

Note that, for both PTB-XL ECG and NIH-Chest X-Ray datasets, FairFed/FR baseline will not be included in the results. The Fair Linear Representation (FR) approach is based on correlations between data features and sensitive attributes and, to the best of our knowledge, does not apply to signal or image datasets~\cite{yuzi_he_geometric_2020}.
Therefore, we will only use FairFed and FedAvg as our baselines for the following experiments. On the other hand, as promised in Section~\ref{sec:exp_heter}, now we will also include the results for final-layer fine-tuning method (FT) with our framework in addition to results with the PP method.

\paragraph{Model Output Post-Processing (PP)}
Our framework with model output post-processing (PP) for fairness is still very effective on more complex datasets and models and outperforms all the baselines in fairness improvement. We can see from Table \ref{tab:stats_ptb} that the effectiveness of PP is not influenced much by the change of data heterogeneity, with an improvement of EOD by 87\%, 83\% and 86\% at $\alpha=0.5$, $\alpha=5$, $\alpha=500$, respectively. On the other hand, the improvement in fairness comes with a decrease of balanced accuracy by 13\% - 16\% across different heterogeneity settings, with the most balanced data distribution having the largest accuracy decrease.
One possible reason could be that on the dataset with a higher heterogeneous distribution, local post-processing could fit the local distribution better.
As for training time and communication costs, the values for PP are still as low as those for the FedAvg baseline

\paragraph{Final-layer fine-tuning (FT)}
Like with PP, our framework with final-layer fine tuning (FT) also outperforms all the baselines in weighted averaged fairness improvement over all heterogeneity settings, even though it does not provide as satisfactory EOD reduction as PP.
Final layer fine-tuning (FT) works better with more heterogeneous data, providing a fairness improvement of 12\% with $\alpha=0.5$, and a slightly lower improvement of 7\% with  $\alpha=500$.
However, it is encouraging to note that FT also provides an increase of BA across all heterogeneity settings while reducing EOD.
Compared with FairFed baselines, FT succeeds both in fairness and in balanced accuracy (BA), with a more efficient training process as measured with time and network communication. While FT overall does not appear satisfactory in our experiments for fairness, we leave it for future work to explore whether different hyper-parameter setting (e.g. for $\alpha_{ft}$, lr, etc) can produce better results.

\subsection{Performance on NIH Chest X-Ray Dataset}
\label{sec:nih}

\begin{table}[htbp]
\centering
\begin{tabular}{c | c | c | c c c c} 
 % \hline
 % \hline
 \multicolumn{7}{c}{\textbf{NIH-Chest X-Ray Dataset}} \\ [0.5ex] 
 \hline
 \hline 
  % \multirow{2}{*}{\makecell{Hetero.\\Level $\alpha$}}
  \multirow{2}{*}{\makecell{$\alpha$}}
  & \multirow{2}{*}{Metric} & \multirow{2}{*}{Client} &  \multicolumn{4}{c}{Method}\\
 \cline{4-7}
 & &
 & FedAvg & FairFed & PP & FT \\
 \hline
 \hline
  % \multirow{11}{*}{$\alpha=0.5$}
  \multirow{11}{*}{$0.5$} 
  & \multirow{5}{*}{BA} 
  &  Avg & 0.846 & 0.840 & {0.844} & \cellcolor[gray]{0.9}\textbf{0.850}\\
  && C1 & \cellcolor[gray]{0.9}\textbf{0.916} & {0.900} & 0.915 & {0.911}\\
  && C2 & 0.577 & \cellcolor[gray]{0.9}\textbf{0.584} &  0.573 & 0.574\\
  && C3 & \cellcolor[gray]{0.9}\textbf{0.962} & 0.961 & {0.960} &  {0.960}\\
  && C4 & {0.888} & 0.870 & 0.885 & \cellcolor[gray]{0.9}\textbf{0.921}\\
  \hhline{~|*{6}{-}}
  & \multirow{5}{*}{EOD} 
  & Avg & 0.061 & \cellcolor[gray]{0.9}\textbf{0.031} & {0.039} & {0.037} \\
  && C1 & 0.106 & \cellcolor[gray]{0.9}\textbf{0.040} & {0.090} & {0.086}\\
  && C2 & 0.013 & 0.008 & {0.011} & \cellcolor[gray]{0.9}\textbf{0.004} \\
  && C3 & 0.044 & 0.035 & {0.044} &  \cellcolor[gray]{0.9}\textbf{0.007}\\
  && C4 & 0.098 & 0.042 & \cellcolor[gray]{0.9}\textbf{0.010} & {0.075} \\
  \hhline{~|*{6}{-}}
  & \multicolumn{2}{c|}{Time} 
  & \cellcolor[gray]{0.9} \textbf{18 min} & 25 min & \cellcolor[gray]{0.9}\textbf{18 min} & 22 min\\
  \hhline{~|*{6}{-}}
  & \multicolumn{2}{c|}{\makecell{Comm. Rounds}} 
   &  \cellcolor[gray]{0.9} \textbf{52} & \cellcolor[gray]{0.9} \textbf{52} &  \cellcolor[gray]{0.9} \textbf{52} &  \cellcolor[gray]{0.9} \textbf{52} \\
  \hline
  \hline
  % \multirow{11}{*}{$\alpha=5$} 
  \multirow{11}{*}{$5$} 
  & \multirow{5}{*}{BA} 
  & Avg & \cellcolor[gray]{0.9}\textbf{0.867} & 0.862  & {0.864} & 0.865\\
  && C1 & \cellcolor[gray]{0.9}\textbf{0.844} & 0.839 & {0.839} & \cellcolor[gray]{0.9}\textbf{0.844} \\
  && C2 & 0.892 & \cellcolor[gray]{0.9}\textbf{0.899} & 0.893 & 0.889\\
  && C3 & \cellcolor[gray]{0.9}\textbf{0.836} & 0.831 & {0.832} & 0.830 \\
  && C4 & 0.890 & 0.871 & {0.887} & \cellcolor[gray]{0.9}\textbf{0.892}\\
  \hhline{~|*{6}{-}}
  & \multirow{5}{*}{EOD} 
  & Avg & 0.051 & \cellcolor[gray]{0.9}\textbf{0.031} & {0.045} & 0.042 \\
  && C1 & 0.035 & 0.061 & \cellcolor[gray]{0.9}\textbf{0.015} & 0.026\\
  && C2 & 0.056 & \cellcolor[gray]{0.9}\textbf{0.019} & {0.080} & 0.041\\
  && C3 & 0.067 & \cellcolor[gray]{0.9}\textbf{0.023} & {0.049} & 0.067\\
  && C4 & 0.046 & \cellcolor[gray]{0.9}\textbf{0.025} & {0.029} & 0.031\\
  \hhline{~|*{6}{-}}
  & \multicolumn{2}{c|}{Time}  
  & \cellcolor[gray]{0.9} \textbf{18 min} & 25 min & \cellcolor[gray]{0.9}\textbf{18 min} & 20 min\\
  \hhline{~|*{6}{-}}
  & \multicolumn{2}{c|}{\makecell{Comm. Rounds}} 
&  \cellcolor[gray]{0.9} \textbf{52} & \cellcolor[gray]{0.9} \textbf{52} &  \cellcolor[gray]{0.9} \textbf{52} &  \cellcolor[gray]{0.9} \textbf{52} \\
  \hline
  \hline
  % \multirow{11}{*}{$\alpha=500$} 
  \multirow{11}{*}{$500$} 
  & \multirow{5}{*}{BA} 
  & Avg & \cellcolor[gray]{0.9}\textbf{0.865} & \cellcolor[gray]{0.9}\textbf{0.865} & 0.862 & {0.864} \\
  && C1 & {0.860} & \cellcolor[gray]{0.9}\textbf{0.861} & 0.857 & {0.857} \\
  && C2 & \cellcolor[gray]{0.9}\textbf{0.879} & {0.873} & 0.877 & {0.878} \\
  && C3 & {0.860} & 0.861 & {0.858} & \cellcolor[gray]{0.9}\textbf{0.865} \\
  && C4 & {0.860} & \cellcolor[gray]{0.9}\textbf{0.864} & {0.856} & 0.859 \\
  \hhline{~|*{6}{-}}
  & \multirow{5}{*}{EOD} 
  & Avg & 0.067 & \cellcolor[gray]{0.9}\textbf{0.024} & {0.049}  & 0.061 \\
  && C1 & 0.103 & \cellcolor[gray]{0.9}\textbf{0.038} & {0.069} & 0.104 \\
  && C2 & 0.035 & 0.007 & \cellcolor[gray]{0.9}\textbf{0.006} & \cellcolor[gray]{0.9}\textbf{0.006} \\
  && C3 & 0.098 & \cellcolor[gray]{0.9}\textbf{0.011} & {0.098} & 0.104 \\
  && C4 & 0.035 & 0.039 & \cellcolor[gray]{0.9}\textbf{0.024} & 0.036 \\
  \hhline{~|*{6}{-}}
  & \multicolumn{2}{c|}{Time}    & \cellcolor[gray]{0.9} \textbf{18 min} & 25 min & \cellcolor[gray]{0.9}\textbf{18 min} & 20 min\\
   \hhline{~|*{6}{-}}
  & \multicolumn{2}{c|}{\makecell{Comm. Rounds}} 
 &  \cellcolor[gray]{0.9} \textbf{52} & \cellcolor[gray]{0.9} \textbf{52} &  \cellcolor[gray]{0.9} \textbf{52} &  \cellcolor[gray]{0.9} \textbf{52} \\
  \hline
  \hline
\end{tabular}
\caption{Client-wise performance comparison on NIH Chest X-Ray dataset with different data heterogeneity level $\alpha$.
BA: Balanced accuracy (Higher is better). EOD: Equalized odds difference (Lower is better). 
Grey highlights indicate the best performance of each row.
FedAvg and FairFed are baselines described in Section \ref{sec:baselines}. 
PP and FT refer to our framework with output post-processing and final-layer fine-tuning respectively.}
\label{tab:stats_nih}
\end{table}

Table \ref{tab:stats_nih} shows the experiment results on the NIH Chest X-Ray dataset for different heterogeneity settings.

In general, our framework with both PP and FT still provide a fairness improvement across all heterogeneity settings which decreases with the decrease of data heterogeneity, from around 36\% to 26\% and from 39\% to 8\% respectively.
And the accuracy decrease is small as well.
FairFed provides a higher fairness EOD improvement on average than our methods while providing a slightly lower accuracy.

Compared with the results on other datasets we discussed so far, we can see that the best performance of different metrics are more unevenly distributed across different methods. For example, looking at weighted average fairness, baseline FairFed appears to perform better. For client-wise fairness metrics, client 2 can achieve the lowest EOD with FT for $\alpha=0.5$ and $\alpha=500$ while client 4 can achieve that with PP instead. Thus, the winner trend is not as clear as it was in the case of previous datasets.

The difference between results on the NIH-Chest X-Ray dataset and the other datasets could be because the hyper-parameters are more sensitive on NIH Chest X-Ray dataset and thus requires more careful tuning and a larger search space to obtain the optimal model. This sensitivity may be due to many reasons.
Firstly, we see from the tables that the EOD metrics on NIH Chest X-Ray obtained by FedAvg training without any fairness constraints are already very small.
Recall that EOD of FedAvg on COMPAS and PTB-XL is generally larger than 0.3, on Adult is more than 0.16, while on NIH Chest X-Ray is around 0.05 - 0.06. Secondly, recall from Table \ref{tab:nih} that NIH Chest X-Ray dataset is very imbalanced with only 18\% of the samples having positive labels (``Effusion'').
After data is split across clients, the local sample size for certain groups could be very small which makes it more difficult to make fine-tuning work. These results highlight some of the limitations of the proposed framework that we will summarize as we approach the end of our paper.

\subsection{Summary of Observed Strengths and Limitations}
\textbf{Model Output Post-Processing: } In the analyses of our framework with PP (i.e. model output post-processing) method, we observed several of its strengths.
Firstly, this method shows its effectiveness with significant fairness improvement across all datasets under various heterogeneity settings. It is especially effective when the original model or data contains relatively large EOD or when local datasets are highly heterogeneous. Secondly, this method is very efficient to apply because it is very fast to compute the derived predictor and requires no hyperparameter tuning. Additionally, it requires minimal computational resources on local clients, making it a low-cost solution (i.e. no GPUs required). Thirdly, the derived predictor generated by PP is human-interpretable (i.e. clients can see how predictions are changed for fairness). On the negative side, we observed that PP comes with drop in accuracy in nearly all cases. While a drop in accuracy is commonly observed in the fairness literature and may often be due to data characteristics and assumptions~\cite{dutta2020there}, but it also shows the importance of considering application and context in employing any fairness intervention. As we discussed in Section~\ref{sec:data-datasets}, application grounded discussions with ethical, legal, domain experts and various stakeholders must be taken into account to select the right fairness intervention. 

\smallskip
\noindent {\textbf{Final Layer Fine-Tuning: }} In the analysis of our framework with FT (i.e. fair final layer fine-tuning) method, we found it to be also effective in improving model fairness. It improves the EOD under all heterogeneity settings for most datasets. Similar to PP, this approach tends to achieve better fairness improvement with more heterogeneous data across clients. In contrast to PP, fine-tuning also provides a better model accuracy under more heterogeneous settings on most datasets. Notably, in cases of extreme data heterogeneity, fine-tuning often improves both fairness and accuracy simultaneously.
While it does not always deliver as large a fairness boost as PP, it delivers fairness improvements with minimal or no impact on model performance.

While not as computationally inexpensive as PP, FT is still efficient because only the last layer of the model is updated, while the other layers are kept fixed. This ensures that even with large models like the deep neural network (DNN) we used in the NIH Chest X-Ray experiments, the time required in this method is still better compared to other fairness baselines. Moreover, since the fine-tuning procedure is performed fully locally, there are no additional communication costs. Local clients can also tune their local model to suit their specific fairness requirements by tweaking parameters such as $\alpha_{ft}$ and the number of local training rounds. This provides a more flexible fairness option for clients compared to the PP approach.

However, this dependence on the fine-tuning parameter $\alpha_{ft}$ also means that achieving the optimal performance requires tuning the hyperparameters, which adds to the cost and complexity of applying it, especially on large datasets with increasing complexity. Finally, in cases where local datasets are small and thus, there are insufficient samples for certain labels/groups, fine-tuning tends to not work satisfactorily and hyperparameter selection may be even more difficult.

\section{Conclusions and Future Work}
In this work, we formally defined a simple post-processing-based fair federated learning (FL) framework, aiming to improve group fairness for each of the clients.
This framework relies on well-established FL training procedure and fairness post-processing approaches commonly used in centralized machine learning, allowing each client to independently apply fairness methods on their local data. We demonstrated the framework with two different fairness post-processing techniques: model output post-processing and final layer fine-tuning. Through comprehensive experiments on four different datasets (tabular, ECG, X-Ray) and with varying degrees of client data heterogeneity, we analyze the strengths and limitations of this framework. The framework decentralizes fairness enforcement by providing the clients with a computationally efficient way of obtaining fairer FL models with the flexibility of choosing different fairness definitions and requirements guided by local context and application needs.

Future work can consider other group fairness definitions that were not covered in our work and individual fairness~\cite{dwork_fairness_2011}. Similarly, settings beyond binary classification can be considered in future work (e.g. non-binary classification tasks or bias in large generative AI models etc). In our experiments, we only considered heterogeneity across clients due to their local datasets. This can be supplemented in future work by considering other kinds of differences (for e.g. different clients optimally selecting hyper-parameters for final layer fine-tuning method for their local context and datasets).

\bibliographystyle{ACM-Reference-Format}
\bibliography{references}


\begin{thebibliography}{56}

%%% the Web address.

\ifx \showCODEN    \undefined \def \showCODEN     #1{\unskip}     \fi
\ifx \showDOI      \undefined \def \showDOI       #1{#1}\fi
\ifx \showISBNx    \undefined \def \showISBNx     #1{\unskip}     \fi
\ifx \showISBNxiii \undefined \def \showISBNxiii  #1{\unskip}     \fi
\ifx \showISSN     \undefined \def \showISSN      #1{\unskip}     \fi
\ifx \showLCCN     \undefined \def \showLCCN      #1{\unskip}     \fi
\ifx \shownote     \undefined \def \shownote      #1{#1}          \fi
\ifx \showarticletitle \undefined \def \showarticletitle #1{#1}   \fi
\ifx \showURL      \undefined \def \showURL       {\relax}        \fi
% invisible to TeX
\providecommand\bibfield[2]{#2}
\providecommand\bibinfo[2]{#2}
\providecommand\natexlab[1]{#1}
\providecommand\showeprint[2][]{arXiv:#2}

\bibitem[noa(2024)]%
        {noauthor_python_2024}
 \bibinfo{year}{2024}\natexlab{}.
\newblock \bibinfo{title}{Python {Imaging} {Library}}.
\newblock
\newblock
\urldef\tempurl%
\url{https://en.wikipedia.org/w/index.php?title=Python_Imaging_Library&oldid=1225854061}
\showURL{%
\tempurl}
\newblock
\shownote{Page Version ID: 1225854061}.


\bibitem[Abay et~al\mbox{.}(2020)]%
        {abay_mitigating_2020}
\bibfield{author}{\bibinfo{person}{Annie Abay}, \bibinfo{person}{Yi Zhou},
  \bibinfo{person}{Nathalie Baracaldo}, \bibinfo{person}{Shashank Rajamoni},
  \bibinfo{person}{Ebube Chuba}, {and} \bibinfo{person}{Heiko Ludwig}.}
  \bibinfo{year}{2020}\natexlab{}.
\newblock \bibinfo{title}{Mitigating {Bias} in {Federated} {Learning}}.
\newblock
\newblock
\urldef\tempurl%
\url{https://doi.org/10.48550/arXiv.2012.02447}
\showDOI{\tempurl}
\newblock
\shownote{arXiv:2012.02447 [cs, stat]}.


\bibitem[Acar et~al\mbox{.}(2021)]%
        {acar_debiasing_2021}
\bibfield{author}{\bibinfo{person}{Durmus Alp~Emre Acar}, \bibinfo{person}{Yue
  Zhao}, \bibinfo{person}{Ruizhao Zhu}, \bibinfo{person}{Ramon Matas},
  \bibinfo{person}{Matthew Mattina}, \bibinfo{person}{Paul Whatmough}, {and}
  \bibinfo{person}{Venkatesh Saligrama}.} \bibinfo{year}{2021}\natexlab{}.
\newblock \showarticletitle{Debiasing {Model} {Updates} for {Improving}
  {Personalized} {Federated} {Training}}. In
  \bibinfo{booktitle}{\emph{Proceedings of the 38th {International}
  {Conference} on {Machine} {Learning}}}. \bibinfo{publisher}{PMLR},
  \bibinfo{pages}{21--31}.
\newblock
\urldef\tempurl%
\url{https://proceedings.mlr.press/v139/acar21a.html}
\showURL{%
\tempurl}
\newblock
\shownote{ISSN: 2640-3498}.


\bibitem[Agarap(2019)]%
        {agarap_deep_2019}
\bibfield{author}{\bibinfo{person}{Abien~Fred Agarap}.}
  \bibinfo{year}{2019}\natexlab{}.
\newblock \bibinfo{title}{Deep {Learning} using {Rectified} {Linear} {Units}
  ({ReLU})}.
\newblock
\newblock
\urldef\tempurl%
\url{https://doi.org/10.48550/arXiv.1803.08375}
\showDOI{\tempurl}
\newblock
\shownote{arXiv:1803.08375 [cs, stat]}.


\bibitem[Barocas et~al\mbox{.}(2023)]%
        {barocas2023fairness}
\bibfield{author}{\bibinfo{person}{Solon Barocas}, \bibinfo{person}{Moritz
  Hardt}, {and} \bibinfo{person}{Arvind Narayanan}.}
  \bibinfo{year}{2023}\natexlab{}.
\newblock \bibinfo{booktitle}{\emph{Fairness and machine learning: Limitations
  and opportunities}}.
\newblock \bibinfo{publisher}{MIT press}.
\newblock


\bibitem[Barry~Becker(1996)]%
        {barry_becker_adult_1996}
\bibfield{author}{\bibinfo{person}{Ronny~Kohavi Barry~Becker}.}
  \bibinfo{year}{1996}\natexlab{}.
\newblock \bibinfo{title}{Adult}.
\newblock
\newblock
\urldef\tempurl%
\url{https://doi.org/10.24432/C5XW20}
\showDOI{\tempurl}


\bibitem[Bellamy et~al\mbox{.}(2018)]%
        {bellamy_ai_2018}
\bibfield{author}{\bibinfo{person}{Rachel K.~E. Bellamy},
  \bibinfo{person}{Kuntal Dey}, \bibinfo{person}{Michael Hind},
  \bibinfo{person}{Samuel~C. Hoffman}, \bibinfo{person}{Stephanie Houde},
  \bibinfo{person}{Kalapriya Kannan}, \bibinfo{person}{Pranay Lohia},
  \bibinfo{person}{Jacquelyn Martino}, \bibinfo{person}{Sameep Mehta},
  \bibinfo{person}{Aleksandra Mojsilovic}, \bibinfo{person}{Seema Nagar},
  \bibinfo{person}{Karthikeyan~Natesan Ramamurthy}, \bibinfo{person}{John
  Richards}, \bibinfo{person}{Diptikalyan Saha}, \bibinfo{person}{Prasanna
  Sattigeri}, \bibinfo{person}{Moninder Singh}, \bibinfo{person}{Kush~R.
  Varshney}, {and} \bibinfo{person}{Yunfeng Zhang}.}
  \bibinfo{year}{2018}\natexlab{}.
\newblock \bibinfo{title}{{AI} {Fairness} 360: {An} {Extensible} {Toolkit} for
  {Detecting}, {Understanding}, and {Mitigating} {Unwanted} {Algorithmic}
  {Bias}}.
\newblock
\newblock
\urldef\tempurl%
\url{https://doi.org/10.48550/arXiv.1810.01943}
\showDOI{\tempurl}
\newblock
\shownote{arXiv:1810.01943 [cs]}.


\bibitem[Binns(2020)]%
        {binns_apparent_2020}
\bibfield{author}{\bibinfo{person}{Reuben Binns}.}
  \bibinfo{year}{2020}\natexlab{}.
\newblock \showarticletitle{On the apparent conflict between individual and
  group fairness}. In \bibinfo{booktitle}{\emph{Proceedings of the 2020
  {Conference} on {Fairness}, {Accountability}, and {Transparency}}}
  \emph{(\bibinfo{series}{{FAT}* '20})}. \bibinfo{publisher}{Association for
  Computing Machinery}, \bibinfo{address}{New York, NY, USA},
  \bibinfo{pages}{514--524}.
\newblock
\showISBNx{978-1-4503-6936-7}
\urldef\tempurl%
\url{https://doi.org/10.1145/3351095.3372864}
\showDOI{\tempurl}


\bibitem[Brodersen et~al\mbox{.}(2010)]%
        {brodersen_balanced_2010}
\bibfield{author}{\bibinfo{person}{Kay~Henning Brodersen},
  \bibinfo{person}{Cheng~Soon Ong}, \bibinfo{person}{Klaas~Enno Stephan}, {and}
  \bibinfo{person}{Joachim~M. Buhmann}.} \bibinfo{year}{2010}\natexlab{}.
\newblock \showarticletitle{The {Balanced} {Accuracy} and {Its} {Posterior}
  {Distribution}}. In \bibinfo{booktitle}{\emph{2010 20th {International}
  {Conference} on {Pattern} {Recognition}}}. \bibinfo{pages}{3121--3124}.
\newblock
\urldef\tempurl%
\url{https://doi.org/10.1109/ICPR.2010.764}
\showDOI{\tempurl}
\newblock
\shownote{ISSN: 1051-4651}.


\bibitem[Che et~al\mbox{.}(2024)]%
        {che2024training}
\bibfield{author}{\bibinfo{person}{Xin Che}, \bibinfo{person}{Jingdi Hu},
  \bibinfo{person}{Zirui Zhou}, \bibinfo{person}{Yong Zhang}, {and}
  \bibinfo{person}{Lingyang Chu}.} \bibinfo{year}{2024}\natexlab{}.
\newblock \showarticletitle{Training Fair Models in Federated Learning without
  Data Privacy Infringement}. In \bibinfo{booktitle}{\emph{2024 IEEE
  International Conference on Big Data (BigData)}}. IEEE,
  \bibinfo{pages}{7687--7696}.
\newblock


\bibitem[Du et~al\mbox{.}(2020)]%
        {du_fairness-aware_2020}
\bibfield{author}{\bibinfo{person}{Wei Du}, \bibinfo{person}{Depeng Xu},
  \bibinfo{person}{Xintao Wu}, {and} \bibinfo{person}{Hanghang Tong}.}
  \bibinfo{year}{2020}\natexlab{}.
\newblock \bibinfo{title}{Fairness-aware {Agnostic} {Federated} {Learning}}.
\newblock
\newblock
\urldef\tempurl%
\url{https://doi.org/10.48550/arXiv.2010.05057}
\showDOI{\tempurl}
\newblock
\shownote{arXiv:2010.05057 [cs]}.


\bibitem[Duan et~al\mbox{.}(2024)]%
        {duan2024post}
\bibfield{author}{\bibinfo{person}{Yuying Duan}, \bibinfo{person}{Yijun Tian},
  \bibinfo{person}{Nitesh Chawla}, {and} \bibinfo{person}{Michael Lemmon}.}
  \bibinfo{year}{2024}\natexlab{}.
\newblock \showarticletitle{Post-Fair Federated Learning: Achieving Group and
  Community Fairness in Federated Learning via Post-processing}.
\newblock \bibinfo{journal}{\emph{arXiv preprint arXiv:2405.17782}}
  (\bibinfo{year}{2024}).
\newblock


\bibitem[Dutta et~al\mbox{.}(2020)]%
        {dutta2020there}
\bibfield{author}{\bibinfo{person}{Sanghamitra Dutta}, \bibinfo{person}{Dennis
  Wei}, \bibinfo{person}{Hazar Yueksel}, \bibinfo{person}{Pin-Yu Chen},
  \bibinfo{person}{Sijia Liu}, {and} \bibinfo{person}{Kush Varshney}.}
  \bibinfo{year}{2020}\natexlab{}.
\newblock \showarticletitle{Is there a trade-off between fairness and accuracy?
  a perspective using mismatched hypothesis testing}. In
  \bibinfo{booktitle}{\emph{International conference on machine learning}}.
  PMLR, \bibinfo{pages}{2803--2813}.
\newblock


\bibitem[Dwork et~al\mbox{.}(2011)]%
        {dwork_fairness_2011}
\bibfield{author}{\bibinfo{person}{Cynthia Dwork}, \bibinfo{person}{Moritz
  Hardt}, \bibinfo{person}{Toniann Pitassi}, \bibinfo{person}{Omer Reingold},
  {and} \bibinfo{person}{Rich Zemel}.} \bibinfo{year}{2011}\natexlab{}.
\newblock \bibinfo{title}{Fairness {Through} {Awareness}}.
\newblock
\newblock
\urldef\tempurl%
\url{https://doi.org/10.48550/arXiv.1104.3913}
\showDOI{\tempurl}
\newblock
\shownote{arXiv:1104.3913 [cs]}.


\bibitem[Ezzeldin et~al\mbox{.}(2023)]%
        {ezzeldin_fairfed_2023}
\bibfield{author}{\bibinfo{person}{Yahya~H. Ezzeldin}, \bibinfo{person}{Shen
  Yan}, \bibinfo{person}{Chaoyang He}, \bibinfo{person}{Emilio Ferrara}, {and}
  \bibinfo{person}{A.~Salman Avestimehr}.} \bibinfo{year}{2023}\natexlab{}.
\newblock \showarticletitle{{FairFed}: {Enabling} {Group} {Fairness} in
  {Federated} {Learning}}.
\newblock \bibinfo{journal}{\emph{Proceedings of the AAAI Conference on
  Artificial Intelligence}} \bibinfo{volume}{37}, \bibinfo{number}{6}
  (\bibinfo{date}{June} \bibinfo{year}{2023}), \bibinfo{pages}{7494--7502}.
\newblock
\showISSN{2374-3468}
\urldef\tempurl%
\url{https://doi.org/10.1609/aaai.v37i6.25911}
\showDOI{\tempurl}
\newblock
\shownote{Number: 6}.


\bibitem[Grote and Keeling(2022)]%
        {grote2022enabling}
\bibfield{author}{\bibinfo{person}{Thomas Grote} {and} \bibinfo{person}{Geoff
  Keeling}.} \bibinfo{year}{2022}\natexlab{}.
\newblock \showarticletitle{Enabling fairness in healthcare through machine
  learning}.
\newblock \bibinfo{journal}{\emph{Ethics and Information Technology}}
  \bibinfo{volume}{24}, \bibinfo{number}{3} (\bibinfo{year}{2022}),
  \bibinfo{pages}{39}.
\newblock


\bibitem[Hardt et~al\mbox{.}(2016)]%
        {hardt_equality_2016}
\bibfield{author}{\bibinfo{person}{Moritz Hardt}, \bibinfo{person}{Eric Price},
  {and} \bibinfo{person}{Nathan Srebro}.} \bibinfo{year}{2016}\natexlab{}.
\newblock \bibinfo{title}{Equality of {Opportunity} in {Supervised}
  {Learning}}.
\newblock
\newblock
\urldef\tempurl%
\url{https://doi.org/10.48550/arXiv.1610.02413}
\showDOI{\tempurl}
\newblock
\shownote{arXiv:1610.02413 [cs]}.


\bibitem[He et~al\mbox{.}(2016)]%
        {he_identity_2016}
\bibfield{author}{\bibinfo{person}{Kaiming He}, \bibinfo{person}{Xiangyu
  Zhang}, \bibinfo{person}{Shaoqing Ren}, {and} \bibinfo{person}{Jian Sun}.}
  \bibinfo{year}{2016}\natexlab{}.
\newblock \bibinfo{title}{Identity {Mappings} in {Deep} {Residual} {Networks}}.
\newblock
\newblock
\urldef\tempurl%
\url{https://doi.org/10.48550/arXiv.1603.05027}
\showDOI{\tempurl}
\newblock
\shownote{arXiv:1603.05027 [cs]}.


\bibitem[Holste et~al\mbox{.}(2022)]%
        {holste_long-tailed_2022}
\bibfield{author}{\bibinfo{person}{Gregory Holste}, \bibinfo{person}{Song
  Wang}, \bibinfo{person}{Ziyu Jiang}, \bibinfo{person}{Thomas~C. Shen},
  \bibinfo{person}{George Shih}, \bibinfo{person}{Ronald~M. Summers},
  \bibinfo{person}{Yifan Peng}, {and} \bibinfo{person}{Zhangyang Wang}.}
  \bibinfo{year}{2022}\natexlab{}.
\newblock \showarticletitle{Long-{Tailed} {Classification} of {Thorax}
  {Diseases} on {Chest} {X}-{Ray}: {A} {New} {Benchmark} {Study}}. In
  \bibinfo{booktitle}{\emph{Data {Augmentation}, {Labelling}, and
  {Imperfections}}}, \bibfield{editor}{\bibinfo{person}{Hien~V. Nguyen},
  \bibinfo{person}{Sharon~X. Huang}, {and} \bibinfo{person}{Yuan Xue}} (Eds.).
  \bibinfo{publisher}{Springer Nature Switzerland}, \bibinfo{address}{Cham},
  \bibinfo{pages}{22--32}.
\newblock
\showISBNx{978-3-031-17027-0}
\urldef\tempurl%
\url{https://doi.org/10.1007/978-3-031-17027-0_3}
\showDOI{\tempurl}


\bibitem[Holstein et~al\mbox{.}(2019)]%
        {holstein2019improving}
\bibfield{author}{\bibinfo{person}{Kenneth Holstein}, \bibinfo{person}{Jennifer
  Wortman~Vaughan}, \bibinfo{person}{Hal Daum{\'e}~III}, \bibinfo{person}{Miro
  Dudik}, {and} \bibinfo{person}{Hanna Wallach}.}
  \bibinfo{year}{2019}\natexlab{}.
\newblock \showarticletitle{Improving fairness in machine learning systems:
  What do industry practitioners need?}. In
  \bibinfo{booktitle}{\emph{Proceedings of the 2019 CHI conference on human
  factors in computing systems}}. \bibinfo{pages}{1--16}.
\newblock


\bibitem[Ioffe and Szegedy(2015)]%
        {ioffe_batch_2015}
\bibfield{author}{\bibinfo{person}{Sergey Ioffe} {and}
  \bibinfo{person}{Christian Szegedy}.} \bibinfo{year}{2015}\natexlab{}.
\newblock \bibinfo{title}{Batch {Normalization}: {Accelerating} {Deep}
  {Network} {Training} by {Reducing} {Internal} {Covariate} {Shift}}.
\newblock
\newblock
\urldef\tempurl%
\url{https://doi.org/10.48550/arXiv.1502.03167}
\showDOI{\tempurl}
\newblock
\shownote{arXiv:1502.03167 [cs]}.


\bibitem[Kingma and Ba(2017)]%
        {kingma_adam_2017}
\bibfield{author}{\bibinfo{person}{Diederik~P. Kingma} {and}
  \bibinfo{person}{Jimmy Ba}.} \bibinfo{year}{2017}\natexlab{}.
\newblock \bibinfo{title}{Adam: {A} {Method} for {Stochastic} {Optimization}}.
\newblock
\newblock
\urldef\tempurl%
\url{https://doi.org/10.48550/arXiv.1412.6980}
\showDOI{\tempurl}
\newblock
\shownote{arXiv:1412.6980 [cs]}.


\bibitem[Larrazabal et~al\mbox{.}(2020)]%
        {larrazabal2020gender}
\bibfield{author}{\bibinfo{person}{Agostina~J Larrazabal},
  \bibinfo{person}{Nicol{\'a}s Nieto}, \bibinfo{person}{Victoria Peterson},
  \bibinfo{person}{Diego~H Milone}, {and} \bibinfo{person}{Enzo Ferrante}.}
  \bibinfo{year}{2020}\natexlab{}.
\newblock \showarticletitle{Gender imbalance in medical imaging datasets
  produces biased classifiers for computer-aided diagnosis}.
\newblock \bibinfo{journal}{\emph{Proceedings of the National Academy of
  Sciences}} \bibinfo{volume}{117}, \bibinfo{number}{23}
  (\bibinfo{year}{2020}), \bibinfo{pages}{12592--12594}.
\newblock


\bibitem[Li et~al\mbox{.}(2020a)]%
        {li2020review}
\bibfield{author}{\bibinfo{person}{Li Li}, \bibinfo{person}{Yuxi Fan},
  \bibinfo{person}{Mike Tse}, {and} \bibinfo{person}{Kuo-Yi Lin}.}
  \bibinfo{year}{2020}\natexlab{a}.
\newblock \showarticletitle{A review of applications in federated learning}.
\newblock \bibinfo{journal}{\emph{Computers \& Industrial Engineering}}
  \bibinfo{volume}{149} (\bibinfo{year}{2020}), \bibinfo{pages}{106854}.
\newblock


\bibitem[Li et~al\mbox{.}(2021)]%
        {li_ditto_2021}
\bibfield{author}{\bibinfo{person}{Tian Li}, \bibinfo{person}{Shengyuan Hu},
  \bibinfo{person}{Ahmad Beirami}, {and} \bibinfo{person}{Virginia Smith}.}
  \bibinfo{year}{2021}\natexlab{}.
\newblock \showarticletitle{Ditto: {Fair} and {Robust} {Federated} {Learning}
  {Through} {Personalization}}. In \bibinfo{booktitle}{\emph{Proceedings of the
  38th {International} {Conference} on {Machine} {Learning}}}.
  \bibinfo{publisher}{PMLR}, \bibinfo{pages}{6357--6368}.
\newblock
\urldef\tempurl%
\url{https://proceedings.mlr.press/v139/li21h.html}
\showURL{%
\tempurl}
\newblock
\shownote{ISSN: 2640-3498}.


\bibitem[Li et~al\mbox{.}(2020b)]%
        {li_fair_2020}
\bibfield{author}{\bibinfo{person}{Tian Li}, \bibinfo{person}{Maziar Sanjabi},
  \bibinfo{person}{Ahmad Beirami}, {and} \bibinfo{person}{Virginia Smith}.}
  \bibinfo{year}{2020}\natexlab{b}.
\newblock \bibinfo{title}{Fair {Resource} {Allocation} in {Federated}
  {Learning}}.
\newblock
\newblock
\urldef\tempurl%
\url{https://doi.org/10.48550/arXiv.1905.10497}
\showDOI{\tempurl}
\newblock
\shownote{arXiv:1905.10497 [cs, stat]}.


\bibitem[Lima et~al\mbox{.}(2021)]%
        {lima_deep_2021}
\bibfield{author}{\bibinfo{person}{Emilly~M. Lima},
  \bibinfo{person}{Antônio~H. Ribeiro}, \bibinfo{person}{Gabriela M.~M.
  Paixão}, \bibinfo{person}{Manoel~Horta Ribeiro}, \bibinfo{person}{Marcelo~M.
  Pinto-Filho}, \bibinfo{person}{Paulo~R. Gomes}, \bibinfo{person}{Derick~M.
  Oliveira}, \bibinfo{person}{Ester~C. Sabino}, \bibinfo{person}{Bruce~B.
  Duncan}, \bibinfo{person}{Luana Giatti}, \bibinfo{person}{Sandhi~M. Barreto},
  \bibinfo{person}{Wagner Meira~Jr}, \bibinfo{person}{Thomas~B. Schön}, {and}
  \bibinfo{person}{Antonio Luiz~P. Ribeiro}.} \bibinfo{year}{2021}\natexlab{}.
\newblock \showarticletitle{Deep neural network-estimated electrocardiographic
  age as a mortality predictor}.
\newblock \bibinfo{journal}{\emph{Nature Communications}} \bibinfo{volume}{12},
  \bibinfo{number}{1} (\bibinfo{date}{Aug.} \bibinfo{year}{2021}),
  \bibinfo{pages}{5117}.
\newblock
\showISSN{2041-1723}
\urldef\tempurl%
\url{https://doi.org/10.1038/s41467-021-25351-7}
\showDOI{\tempurl}
\newblock
\shownote{Publisher: Nature Publishing Group}.


\bibitem[Maior et~al\mbox{.}(2021)]%
        {maior_convolutional_2021}
\bibfield{author}{\bibinfo{person}{Caio B.~S. Maior}, \bibinfo{person}{João
  M.~M. Santana}, \bibinfo{person}{Isis~D. Lins}, {and}
  \bibinfo{person}{Márcio J.~C. Moura}.} \bibinfo{year}{2021}\natexlab{}.
\newblock \showarticletitle{Convolutional neural network model based on
  radiological images to support {COVID}-19 diagnosis: {Evaluating} database
  biases}.
\newblock \bibinfo{journal}{\emph{PLOS ONE}} \bibinfo{volume}{16},
  \bibinfo{number}{3} (\bibinfo{date}{March} \bibinfo{year}{2021}),
  \bibinfo{pages}{e0247839}.
\newblock
\showISSN{1932-6203}
\urldef\tempurl%
\url{https://doi.org/10.1371/journal.pone.0247839}
\showDOI{\tempurl}
\newblock
\shownote{Publisher: Public Library of Science}.


\bibitem[Makhija et~al\mbox{.}(2024)]%
        {makhija2024achieving}
\bibfield{author}{\bibinfo{person}{Disha Makhija}, \bibinfo{person}{Xing Han},
  \bibinfo{person}{Joydeep Ghosh}, {and} \bibinfo{person}{Yejin Kim}.}
  \bibinfo{year}{2024}\natexlab{}.
\newblock \showarticletitle{Achieving fairness across local and global models
  in federated learning}.
\newblock \bibinfo{journal}{\emph{arXiv preprint arXiv:2406.17102}}
  (\bibinfo{year}{2024}).
\newblock


\bibitem[Mao et~al\mbox{.}(2023b)]%
        {mao_cross-entropy_2023}
\bibfield{author}{\bibinfo{person}{Anqi Mao}, \bibinfo{person}{Mehryar Mohri},
  {and} \bibinfo{person}{Yutao Zhong}.} \bibinfo{year}{2023}\natexlab{b}.
\newblock \bibinfo{title}{Cross-{Entropy} {Loss} {Functions}: {Theoretical}
  {Analysis} and {Applications}}.
\newblock
\newblock
\urldef\tempurl%
\url{https://doi.org/10.48550/arXiv.2304.07288}
\showDOI{\tempurl}
\newblock
\shownote{arXiv:2304.07288 [cs, stat]}.


\bibitem[Mao et~al\mbox{.}(2023a)]%
        {mao_last-layer_2023}
\bibfield{author}{\bibinfo{person}{Yuzhen Mao}, \bibinfo{person}{Zhun Deng},
  \bibinfo{person}{Huaxiu Yao}, \bibinfo{person}{Ting Ye},
  \bibinfo{person}{Kenji Kawaguchi}, {and} \bibinfo{person}{James Zou}.}
  \bibinfo{year}{2023}\natexlab{a}.
\newblock \bibinfo{title}{Last-{Layer} {Fairness} {Fine}-tuning is {Simple} and
  {Effective} for {Neural} {Networks}}.
\newblock
\newblock
\urldef\tempurl%
\url{https://doi.org/10.48550/arXiv.2304.03935}
\showDOI{\tempurl}
\newblock
\shownote{arXiv:2304.03935 [cs]}.


\bibitem[Marcinkevics et~al\mbox{.}(2022)]%
        {marcinkevics_debiasing_2022}
\bibfield{author}{\bibinfo{person}{Ricards Marcinkevics}, \bibinfo{person}{Ece
  Ozkan}, {and} \bibinfo{person}{Julia~E. Vogt}.}
  \bibinfo{year}{2022}\natexlab{}.
\newblock \showarticletitle{Debiasing {Deep} {Chest} {X}-{Ray} {Classifiers}
  using {Intra}- and {Post}-processing {Methods}}. In
  \bibinfo{booktitle}{\emph{Proceedings of the 7th {Machine} {Learning} for
  {Healthcare} {Conference}}}. \bibinfo{publisher}{PMLR},
  \bibinfo{pages}{504--536}.
\newblock
\urldef\tempurl%
\url{https://proceedings.mlr.press/v182/marcinkevics22a.html}
\showURL{%
\tempurl}
\newblock
\shownote{ISSN: 2640-3498}.


\bibitem[Mattu et~al\mbox{.}(2016)]%
        {mattu_how_2016}
\bibfield{author}{\bibinfo{person}{Mattu}, \bibinfo{person}{{Jeff Larson}},
  \bibinfo{person}{{Julia Angwin}}, \bibinfo{person}{{Lauren Kirchner}}, {and}
  \bibinfo{person}{{Surya}}.} \bibinfo{year}{2016}\natexlab{}.
\newblock \showarticletitle{How {We} {Analyzed} the {COMPAS} {Recidivism}
  {Algorithm}}.
\newblock \bibinfo{journal}{\emph{ProPublica}} (\bibinfo{year}{2016}).
\newblock
\urldef\tempurl%
\url{https://www.propublica.org/article/how-we-analyzed-the-compas-recidivism-algorithm}
\showURL{%
\tempurl}


\bibitem[McMahan et~al\mbox{.}(2017)]%
        {mcmahan_communication-efficient_2017}
\bibfield{author}{\bibinfo{person}{Brendan McMahan}, \bibinfo{person}{Eider
  Moore}, \bibinfo{person}{Daniel Ramage}, \bibinfo{person}{Seth Hampson},
  {and} \bibinfo{person}{Blaise Aguera~y Arcas}.}
  \bibinfo{year}{2017}\natexlab{}.
\newblock \showarticletitle{Communication-{Efficient} {Learning} of {Deep}
  {Networks} from {Decentralized} {Data}}. In
  \bibinfo{booktitle}{\emph{Proceedings of the 20th {International}
  {Conference} on {Artificial} {Intelligence} and {Statistics}}}.
  \bibinfo{publisher}{PMLR}, \bibinfo{pages}{1273--1282}.
\newblock
\urldef\tempurl%
\url{https://proceedings.mlr.press/v54/mcmahan17a.html}
\showURL{%
\tempurl}
\newblock
\shownote{ISSN: 2640-3498}.


\bibitem[Mehrabi et~al\mbox{.}(2022)]%
        {mehrabi_survey_2022}
\bibfield{author}{\bibinfo{person}{Ninareh Mehrabi}, \bibinfo{person}{Fred
  Morstatter}, \bibinfo{person}{Nripsuta Saxena}, \bibinfo{person}{Kristina
  Lerman}, {and} \bibinfo{person}{Aram Galstyan}.}
  \bibinfo{year}{2022}\natexlab{}.
\newblock \showarticletitle{A {Survey} on {Bias} and {Fairness} in {Machine}
  {Learning}}.
\newblock \bibinfo{journal}{\emph{Comput. Surveys}} \bibinfo{volume}{54},
  \bibinfo{number}{6} (\bibinfo{date}{July} \bibinfo{year}{2022}),
  \bibinfo{pages}{1--35}.
\newblock
\showISSN{0360-0300, 1557-7341}
\urldef\tempurl%
\url{https://doi.org/10.1145/3457607}
\showDOI{\tempurl}


\bibitem[Mendieta et~al\mbox{.}(2022)]%
        {mendieta_local_2022}
\bibfield{author}{\bibinfo{person}{Matias Mendieta},
  \bibinfo{person}{Taojiannan Yang}, \bibinfo{person}{Pu Wang},
  \bibinfo{person}{Minwoo Lee}, \bibinfo{person}{Zhengming Ding}, {and}
  \bibinfo{person}{Chen Chen}.} \bibinfo{year}{2022}\natexlab{}.
\newblock \showarticletitle{Local {Learning} {Matters}: {Rethinking} {Data}
  {Heterogeneity} in {Federated} {Learning}}. \bibinfo{pages}{8397--8406}.
\newblock
\urldef\tempurl%
\url{https://openaccess.thecvf.com/content/CVPR2022/html/Mendieta_Local_Learning_Matters_Rethinking_Data_Heterogeneity_in_Federated_Learning_CVPR_2022_paper.html}
\showURL{%
\tempurl}


\bibitem[Mohri et~al\mbox{.}(2019)]%
        {mohri_agnostic_2019}
\bibfield{author}{\bibinfo{person}{Mehryar Mohri}, \bibinfo{person}{Gary
  Sivek}, {and} \bibinfo{person}{Ananda~Theertha Suresh}.}
  \bibinfo{year}{2019}\natexlab{}.
\newblock \showarticletitle{Agnostic {Federated} {Learning}}. In
  \bibinfo{booktitle}{\emph{Proceedings of the 36th {International}
  {Conference} on {Machine} {Learning}}}. \bibinfo{publisher}{PMLR},
  \bibinfo{pages}{4615--4625}.
\newblock
\urldef\tempurl%
\url{https://proceedings.mlr.press/v97/mohri19a.html}
\showURL{%
\tempurl}
\newblock
\shownote{ISSN: 2640-3498}.


\bibitem[Noseworthy et~al\mbox{.}(2020)]%
        {noseworthy2020assessing}
\bibfield{author}{\bibinfo{person}{Peter~A Noseworthy},
  \bibinfo{person}{Zachi~I Attia}, \bibinfo{person}{LaPrincess~C Brewer},
  \bibinfo{person}{Sharonne~N Hayes}, \bibinfo{person}{Xiaoxi Yao},
  \bibinfo{person}{Suraj Kapa}, \bibinfo{person}{Paul~A Friedman}, {and}
  \bibinfo{person}{Francisco Lopez-Jimenez}.} \bibinfo{year}{2020}\natexlab{}.
\newblock \showarticletitle{Assessing and mitigating bias in medical artificial
  intelligence: the effects of race and ethnicity on a deep learning model for
  ECG analysis}.
\newblock \bibinfo{journal}{\emph{Circulation: Arrhythmia and
  Electrophysiology}} \bibinfo{volume}{13}, \bibinfo{number}{3}
  (\bibinfo{year}{2020}), \bibinfo{pages}{e007988}.
\newblock


\bibitem[Paszke et~al\mbox{.}(2019)]%
        {paszke_pytorch_2019}
\bibfield{author}{\bibinfo{person}{Adam Paszke}, \bibinfo{person}{Sam Gross},
  \bibinfo{person}{Francisco Massa}, \bibinfo{person}{Adam Lerer},
  \bibinfo{person}{James Bradbury}, \bibinfo{person}{Gregory Chanan},
  \bibinfo{person}{Trevor Killeen}, \bibinfo{person}{Zeming Lin},
  \bibinfo{person}{Natalia Gimelshein}, \bibinfo{person}{Luca Antiga},
  \bibinfo{person}{Alban Desmaison}, \bibinfo{person}{Andreas Köpf},
  \bibinfo{person}{Edward Yang}, \bibinfo{person}{Zach DeVito},
  \bibinfo{person}{Martin Raison}, \bibinfo{person}{Alykhan Tejani},
  \bibinfo{person}{Sasank Chilamkurthy}, \bibinfo{person}{Benoit Steiner},
  \bibinfo{person}{Lu Fang}, \bibinfo{person}{Junjie Bai}, {and}
  \bibinfo{person}{Soumith Chintala}.} \bibinfo{year}{2019}\natexlab{}.
\newblock \bibinfo{title}{{PyTorch}: {An} {Imperative} {Style},
  {High}-{Performance} {Deep} {Learning} {Library}}.
\newblock
\newblock
\urldef\tempurl%
\url{https://doi.org/10.48550/arXiv.1912.01703}
\showDOI{\tempurl}
\newblock
\shownote{arXiv:1912.01703 [cs, stat]}.


\bibitem[Pedregosa et~al\mbox{.}(2011)]%
        {pedregosa_scikit-learn_2011}
\bibfield{author}{\bibinfo{person}{Fabian Pedregosa}, \bibinfo{person}{Gaël
  Varoquaux}, \bibinfo{person}{Alexandre Gramfort}, \bibinfo{person}{Vincent
  Michel}, \bibinfo{person}{Bertrand Thirion}, \bibinfo{person}{Olivier
  Grisel}, \bibinfo{person}{Mathieu Blondel}, \bibinfo{person}{Peter
  Prettenhofer}, \bibinfo{person}{Ron Weiss}, \bibinfo{person}{Vincent
  Dubourg}, \bibinfo{person}{Jake Vanderplas}, \bibinfo{person}{Alexandre
  Passos}, \bibinfo{person}{David Cournapeau}, \bibinfo{person}{Matthieu
  Brucher}, \bibinfo{person}{Matthieu Perrot}, {and} \bibinfo{person}{Édouard
  Duchesnay}.} \bibinfo{year}{2011}\natexlab{}.
\newblock \showarticletitle{Scikit-learn: {Machine} {Learning} in {Python}}.
\newblock \bibinfo{journal}{\emph{Journal of Machine Learning Research}}
  \bibinfo{volume}{12}, \bibinfo{number}{85} (\bibinfo{year}{2011}),
  \bibinfo{pages}{2825--2830}.
\newblock
\showISSN{1533-7928}
\urldef\tempurl%
\url{http://jmlr.org/papers/v12/pedregosa11a.html}
\showURL{%
\tempurl}


\bibitem[Reshan et~al\mbox{.}(2023)]%
        {reshan_detection_2023}
\bibfield{author}{\bibinfo{person}{Mana Saleh~Al Reshan},
  \bibinfo{person}{Kanwarpartap~Singh Gill}, \bibinfo{person}{Vatsala Anand},
  \bibinfo{person}{Sheifali Gupta}, \bibinfo{person}{Hani Alshahrani},
  \bibinfo{person}{Adel Sulaiman}, {and} \bibinfo{person}{Asadullah Shaikh}.}
  \bibinfo{year}{2023}\natexlab{}.
\newblock \showarticletitle{Detection of {Pneumonia} from {Chest} {X}-ray
  {Images} {Utilizing} {MobileNet} {Model}}.
\newblock \bibinfo{journal}{\emph{Healthcare}} \bibinfo{volume}{11},
  \bibinfo{number}{11} (\bibinfo{date}{Jan.} \bibinfo{year}{2023}),
  \bibinfo{pages}{1561}.
\newblock
\showISSN{2227-9032}
\urldef\tempurl%
\url{https://doi.org/10.3390/healthcare11111561}
\showDOI{\tempurl}
\newblock
\shownote{Number: 11 Publisher: Multidisciplinary Digital Publishing
  Institute}.


\bibitem[Ruder(2017)]%
        {ruder_overview_2017}
\bibfield{author}{\bibinfo{person}{Sebastian Ruder}.}
  \bibinfo{year}{2017}\natexlab{}.
\newblock \bibinfo{title}{An overview of gradient descent optimization
  algorithms}.
\newblock
\newblock
\urldef\tempurl%
\url{https://doi.org/10.48550/arXiv.1609.04747}
\showDOI{\tempurl}
\newblock
\shownote{arXiv:1609.04747 [cs]}.


\bibitem[Salazar et~al\mbox{.}(2024)]%
        {salazar2024survey}
\bibfield{author}{\bibinfo{person}{Teresa Salazar}, \bibinfo{person}{Helder
  Ara{\'u}jo}, \bibinfo{person}{Alberto Cano}, {and}
  \bibinfo{person}{Pedro~Henriques Abreu}.} \bibinfo{year}{2024}\natexlab{}.
\newblock \showarticletitle{A Survey on Group Fairness in Federated Learning:
  Challenges, Taxonomy of Solutions and Directions for Future Research}.
\newblock \bibinfo{journal}{\emph{arXiv preprint arXiv:2410.03855}}
  (\bibinfo{year}{2024}).
\newblock


\bibitem[Sandler et~al\mbox{.}(2019)]%
        {sandler_mobilenetv2_2019}
\bibfield{author}{\bibinfo{person}{Mark Sandler}, \bibinfo{person}{Andrew
  Howard}, \bibinfo{person}{Menglong Zhu}, \bibinfo{person}{Andrey Zhmoginov},
  {and} \bibinfo{person}{Liang-Chieh Chen}.} \bibinfo{year}{2019}\natexlab{}.
\newblock \bibinfo{title}{{MobileNetV2}: {Inverted} {Residuals} and {Linear}
  {Bottlenecks}}.
\newblock
\newblock
\urldef\tempurl%
\url{https://doi.org/10.48550/arXiv.1801.04381}
\showDOI{\tempurl}
\newblock
\shownote{arXiv:1801.04381 [cs]}.


\bibitem[Shamsian et~al\mbox{.}(2021)]%
        {shamsian_personalized_2021}
\bibfield{author}{\bibinfo{person}{Aviv Shamsian}, \bibinfo{person}{Aviv
  Navon}, \bibinfo{person}{Ethan Fetaya}, {and} \bibinfo{person}{Gal Chechik}.}
  \bibinfo{year}{2021}\natexlab{}.
\newblock \showarticletitle{Personalized {Federated} {Learning} using
  {Hypernetworks}}. In \bibinfo{booktitle}{\emph{Proceedings of the 38th
  {International} {Conference} on {Machine} {Learning}}}.
  \bibinfo{publisher}{PMLR}, \bibinfo{pages}{9489--9502}.
\newblock
\urldef\tempurl%
\url{https://proceedings.mlr.press/v139/shamsian21a.html}
\showURL{%
\tempurl}
\newblock
\shownote{ISSN: 2640-3498}.


\bibitem[Srivastava et~al\mbox{.}(2014)]%
        {srivastava_dropout_2014}
\bibfield{author}{\bibinfo{person}{Nitish Srivastava},
  \bibinfo{person}{Geoffrey Hinton}, \bibinfo{person}{Alex Krizhevsky},
  \bibinfo{person}{Ilya Sutskever}, {and} \bibinfo{person}{Ruslan
  Salakhutdinov}.} \bibinfo{year}{2014}\natexlab{}.
\newblock \showarticletitle{Dropout: {A} {Simple} {Way} to {Prevent} {Neural}
  {Networks} from {Overfitting}}.
\newblock \bibinfo{journal}{\emph{Journal of Machine Learning Research}}
  \bibinfo{volume}{15}, \bibinfo{number}{56} (\bibinfo{year}{2014}),
  \bibinfo{pages}{1929--1958}.
\newblock
\showISSN{1533-7928}
\urldef\tempurl%
\url{http://jmlr.org/papers/v15/srivastava14a.html}
\showURL{%
\tempurl}


\bibitem[Wagner et~al\mbox{.}(2022)]%
        {wagner_ptb-xl_2022}
\bibfield{author}{\bibinfo{person}{Patrick Wagner}, \bibinfo{person}{Nils
  Strodthoff}, \bibinfo{person}{Ralf-Dieter Bousseljot},
  \bibinfo{person}{Wojciech Samek}, {and} \bibinfo{person}{Tobias Schaeffter}.}
  \bibinfo{year}{2022}\natexlab{}.
\newblock \bibinfo{title}{{PTB}-{XL}, a large publicly available
  electrocardiography dataset}.
\newblock
\newblock
\urldef\tempurl%
\url{https://doi.org/10.13026/KFZX-AW45}
\showDOI{\tempurl}


\bibitem[Wang et~al\mbox{.}(2024)]%
        {wang_analyzing_2024}
\bibfield{author}{\bibinfo{person}{Tongnian Wang}, \bibinfo{person}{Kai Zhang},
  \bibinfo{person}{Jiannan Cai}, \bibinfo{person}{Yanmin Gong},
  \bibinfo{person}{Kim-Kwang~Raymond Choo}, {and} \bibinfo{person}{Yuanxiong
  Guo}.} \bibinfo{year}{2024}\natexlab{}.
\newblock \showarticletitle{Analyzing the {Impact} of {Personalization} on
  {Fairness} in {Federated} {Learning} for {Healthcare}}.
\newblock \bibinfo{journal}{\emph{Journal of Healthcare Informatics Research}}
  \bibinfo{volume}{8}, \bibinfo{number}{2} (\bibinfo{date}{June}
  \bibinfo{year}{2024}), \bibinfo{pages}{181--205}.
\newblock
\showISSN{2509-498X}
\urldef\tempurl%
\url{https://doi.org/10.1007/s41666-024-00164-7}
\showDOI{\tempurl}


\bibitem[Wang et~al\mbox{.}(2017)]%
        {wang_chestx-ray8_2017}
\bibfield{author}{\bibinfo{person}{Xiaosong Wang}, \bibinfo{person}{Yifan
  Peng}, \bibinfo{person}{Le Lu}, \bibinfo{person}{Zhiyong Lu},
  \bibinfo{person}{Mohammadhadi Bagheri}, {and} \bibinfo{person}{Ronald~M.
  Summers}.} \bibinfo{year}{2017}\natexlab{}.
\newblock \showarticletitle{{ChestX}-ray8: {Hospital}-scale {Chest} {X}-ray
  {Database} and {Benchmarks} on {Weakly}-{Supervised} {Classification} and
  {Localization} of {Common} {Thorax} {Diseases}}. In
  \bibinfo{booktitle}{\emph{2017 {IEEE} {Conference} on {Computer} {Vision} and
  {Pattern} {Recognition} ({CVPR})}}. \bibinfo{pages}{3462--3471}.
\newblock
\urldef\tempurl%
\url{https://doi.org/10.1109/CVPR.2017.369}
\showDOI{\tempurl}
\newblock
\shownote{arXiv:1705.02315 [cs]}.


\bibitem[Yurochkin et~al\mbox{.}(2019)]%
        {yurochkin_bayesian_2019}
\bibfield{author}{\bibinfo{person}{Mikhail Yurochkin}, \bibinfo{person}{Mayank
  Agarwal}, \bibinfo{person}{Soumya Ghosh}, \bibinfo{person}{Kristjan
  Greenewald}, \bibinfo{person}{Trong~Nghia Hoang}, {and}
  \bibinfo{person}{Yasaman Khazaeni}.} \bibinfo{year}{2019}\natexlab{}.
\newblock \bibinfo{title}{Bayesian {Nonparametric} {Federated} {Learning} of
  {Neural} {Networks}}.
\newblock
\newblock
\urldef\tempurl%
\url{https://doi.org/10.48550/arXiv.1905.12022}
\showDOI{\tempurl}
\newblock
\shownote{arXiv:1905.12022 [cs, stat]}.


\bibitem[{Yuzi He} et~al\mbox{.}(2020)]%
        {yuzi_he_geometric_2020}
\bibfield{author}{\bibinfo{person}{{Yuzi He}}, \bibinfo{person}{{Keith
  Burghardt}}, {and} \bibinfo{person}{{Kristina Lerman}}.}
  \bibinfo{year}{2020}\natexlab{}.
\newblock \showarticletitle{A {Geometric} {Solution} to {Fair}
  {Representations}}.
\newblock \bibinfo{journal}{\emph{Proceedings of the AAAI/ACM Conference on AI,
  Ethics, and Society}} (\bibinfo{year}{2020}).
\newblock
\urldef\tempurl%
\url{https://dl.acm.org/doi/abs/10.1145/3375627.3375864}
\showURL{%
\tempurl}


\bibitem[Zafar et~al\mbox{.}(2017)]%
        {zafar_fairness_2017}
\bibfield{author}{\bibinfo{person}{Muhammad~Bilal Zafar},
  \bibinfo{person}{Isabel Valera}, \bibinfo{person}{Manuel~Gomez Rodriguez},
  {and} \bibinfo{person}{Krishna~P. Gummadi}.} \bibinfo{year}{2017}\natexlab{}.
\newblock \showarticletitle{Fairness {Beyond} {Disparate} {Treatment} \&
  {Disparate} {Impact}: {Learning} {Classification} without {Disparate}
  {Mistreatment}}. In \bibinfo{booktitle}{\emph{Proceedings of the 26th
  {International} {Conference} on {World} {Wide} {Web}}}.
  \bibinfo{pages}{1171--1180}.
\newblock
\urldef\tempurl%
\url{https://doi.org/10.1145/3038912.3052660}
\showDOI{\tempurl}
\newblock
\shownote{arXiv:1610.08452 [cs, stat]}.


\bibitem[Zeng et~al\mbox{.}(2022)]%
        {zeng_improving_2022}
\bibfield{author}{\bibinfo{person}{Yuchen Zeng}, \bibinfo{person}{Hongxu Chen},
  {and} \bibinfo{person}{Kangwook Lee}.} \bibinfo{year}{2022}\natexlab{}.
\newblock \bibinfo{title}{Improving {Fairness} via {Federated} {Learning}}.
\newblock
\newblock
\urldef\tempurl%
\url{https://doi.org/10.48550/arXiv.2110.15545}
\showDOI{\tempurl}
\newblock
\shownote{arXiv:2110.15545 [cs]}.


\bibitem[Zhang et~al\mbox{.}(2021)]%
        {zhang2021survey}
\bibfield{author}{\bibinfo{person}{Chen Zhang}, \bibinfo{person}{Yu Xie},
  \bibinfo{person}{Hang Bai}, \bibinfo{person}{Bin Yu},
  \bibinfo{person}{Weihong Li}, {and} \bibinfo{person}{Yuan Gao}.}
  \bibinfo{year}{2021}\natexlab{}.
\newblock \showarticletitle{A survey on federated learning}.
\newblock \bibinfo{journal}{\emph{Knowledge-Based Systems}}
  \bibinfo{volume}{216} (\bibinfo{year}{2021}), \bibinfo{pages}{106775}.
\newblock


\bibitem[Zhang et~al\mbox{.}(2020)]%
        {zhang_fairfl_2020}
\bibfield{author}{\bibinfo{person}{Daniel~Yue Zhang}, \bibinfo{person}{Ziyi
  Kou}, {and} \bibinfo{person}{Dong Wang}.} \bibinfo{year}{2020}\natexlab{}.
\newblock \showarticletitle{{FairFL}: {A} {Fair} {Federated} {Learning}
  {Approach} to {Reducing} {Demographic} {Bias} in {Privacy}-{Sensitive}
  {Classification} {Models}}. In \bibinfo{booktitle}{\emph{2020 {IEEE}
  {International} {Conference} on {Big} {Data} ({Big} {Data})}}.
  \bibinfo{pages}{1051--1060}.
\newblock
\urldef\tempurl%
\url{https://doi.org/10.1109/BigData50022.2020.9378043}
\showDOI{\tempurl}


\bibitem[Zhao et~al\mbox{.}(2024)]%
        {zhao2024federated}
\bibfield{author}{\bibinfo{person}{Joshua~C Zhao}, \bibinfo{person}{Saurabh
  Bagchi}, \bibinfo{person}{Salman Avestimehr}, \bibinfo{person}{Kevin~S Chan},
  \bibinfo{person}{Somali Chaterji}, \bibinfo{person}{Dimitris Dimitriadis},
  \bibinfo{person}{Jiacheng Li}, \bibinfo{person}{Ninghui Li},
  \bibinfo{person}{Arash Nourian}, {and} \bibinfo{person}{Holger~R Roth}.}
  \bibinfo{year}{2024}\natexlab{}.
\newblock \showarticletitle{Federated Learning Privacy: Attacks, Defenses,
  Applications, and Policy Landscape-A Survey}.
\newblock \bibinfo{journal}{\emph{arXiv preprint arXiv:2405.03636}}
  (\bibinfo{year}{2024}).
\newblock


\end{thebibliography}

\appendix

\end{document}